\documentclass[smallcondensed,natbib]{svjour3}    
\smartqed  
\usepackage{graphicx}
\usepackage{mathptmx}      
\usepackage{verbatim} 
\usepackage{hyperref} 
\usepackage[ruled,vlined,linesnumbered,commentsnumbered]{algorithm2e} 
\usepackage{multirow} 
\usepackage{amsmath}  
\usepackage{longtable}
\usepackage{pdflscape}
\usepackage{subcaption} 
\usepackage{multicol} 

\journalname{Machine Learning}

\begin{document}

\title{Multi-Target Regression via Input Space Expansion:\\Treating Targets as Inputs\thanks{The final publication is available at Springer via \url{http://dx.doi.org/10.1007/s10994-016-5546-z.}}}


\author{Eleftherios Spyromitros-Xioufis \and Grigorios Tsoumakas \and William Groves \and Ioannis Vlahavas 
}


\institute{E. Spyromitros-Xioufis \and G. Tsoumakas \and I. Vlahavas
            \at Department of Informatics, Aristotle University of Thessaloniki, Greece \\
              \email{espyromi@csd.auth.gr}, greg@csd.auth.gr, vlahavas@csd.auth.gr
           \and
           W. Groves \at
              Department of Computer Science and Engineering, University of Minnesota, USA \\
              \email{groves@cs.umn.edu}
}

\date{Accepted:15 January 2016}

\maketitle

\begin{abstract}
In many practical applications of supervised learning the task involves the prediction of multiple target variables from a common set of input variables.
When the prediction targets are binary the task is called multi-label classification, while when the targets are continuous the task is called multi-target regression.
In both tasks, target variables often exhibit statistical dependencies and exploiting them in order to improve predictive accuracy is a core challenge.
A family of multi-label classification methods address this challenge by building a separate model for each target on an expanded input space where other targets are treated as additional input variables.
Despite the success of these methods in the multi-label classification domain, their applicability and effectiveness in multi-target regression has not been studied until now.
In this paper, we introduce two new methods for multi-target regression, called \textit{Stacked Single-Target} and \textit{Ensemble of Regressor Chains}, by adapting two popular multi-label classification methods of this family.
Furthermore, we highlight an inherent problem of these methods - a discrepancy of the values of the additional input variables between training and prediction - and develop extensions that use out-of-sample estimates of the target variables during training in order to tackle this problem.
The results of an extensive experimental evaluation carried out on a large and diverse collection of datasets show that, when the discrepancy is appropriately mitigated, the proposed methods attain consistent improvements over the independent regressions baseline.
Moreover, two versions of Ensemble of Regression Chains perform significantly better than four state-of-the-art methods including regularization-based multi-task learning methods and a multi-objective random forest approach.

\end{abstract}

\section{Introduction}
\label{sec:intro}

Multi-target regression (MTR), also known as multivariate or multi-output regression, refers to the task of predicting multiple continuous variables using a common set of input variables.
Such problems arise in various fields including ecological modeling \citep{kocev2009,dzeroski2000} (e.g. predicting the abundance of plant species using water quality measurements), economics \citep{ghosn1997multi} (e.g. predicting stock prices from econometric variables) and energy (e.g. predicting energy production in solar/wind farms using historical measurements and weather forecast information).
Given the importance and diversity of its applications, it is not surprising that research on this topic has started as early as 40 years ago in Statistics \citep{Izenman1975}.

Recently, a closely related task called multi-label classification (MLC) \citep{tsoumakas2010b,zhang2014} has received increased attention by Machine Learning researchers. Similarly to MTR, MLC deals with the prediction of multiple variables using a common set of input variables. However, prediction targets in MLC are binary. 
In fact, the two tasks can be thought of as instances of the more general learning task of multi-target prediction where targets can be continuous, binary, ordinal, categorical or even of mixed type. The baseline approach of learning a separate model for each target applies to both MTR and MLC. Moreover, they share the same core challenge of exploiting dependencies between targets (in addition to dependencies between targets and inputs) in order to improve prediction accuracy, as acknowledged by researchers working in both tasks \citep[e.g.][]{izenman2008,dembczynski2012}. Despite their commonalities, MTR and MLC have typically been treated in isolation and only few works \citep{blockeel1998,weston2002kernel,sieger2005,balasubramanian2012}
have given a general formulation of their key ideas, recognizing the dual applicability of their approaches. 

Motivated by the tight connection between the two tasks, this paper looks at a family of MLC methods that, despite being almost directly applicable to MTR problems, have not been applied so far in this domain. In particular, we consider methods that decompose the MLC task into a series of binary classification tasks, one for each label. This category, includes  the typical one-versus-all or \textit{Binary Relevance} approach that assumes label independence but also approaches that model label dependencies by building models that treat other labels as additional input variables (meta-inputs). 
In this work we adapt two popular methods of this kind \citep{godbole2004,read2011} for MTR, 
contributing two new MTR methods: \textit{Stacked Single-Target} (SST) and \textit{Ensemble of Regressor Chains} (ERC). 
Both methods have been very successful in the MLC domain and provided inspiration for many subsequent works \citep{cheng2009,dembczynski2010a,kumar2012learning,read2014efficient}.

Although the adaptation is trivial (as it basically consists of employing a regression instead of a binary classification algorithm to solve each single-target prediction task), it widens the applicability of existing approaches and increases our understanding of challenges shared by both learning tasks, such as the modeling of target dependencies.
This kind of abstraction of key ideas from solutions tailored to related problems can sometimes offer additional advantages, such as improving the modularity and conceptual simplicity of learning techniques and avoiding reinvention of the same solutions\footnote{See NIPS'11 workshop on relations among machine learning problems at \url{http://rml.anu.edu.au/}}.
 
In addition to evaluating the direct adaptations of the corresponding MLC methods in the MTR domain, we also take a careful look at the treatment of targets as additional input variables and spot a shortcoming that was overlooked in the original MLC formulations of both methods. Specifically, we notice that in both methods the values of the meta-inputs are generated differently between training and prediction, causing a discrepancy that is shown to drastically downgrade their performance.
To tackle this problem, we develop extended versions of the two methods that manage to decrease the discrepancy by using out-of-sample estimates of the targets during training.  These estimates are obtained via an internal cross-validation methodology.

The performance of the proposed methods is comprehensively analyzed based on a large experimental study that includes 18 diverse real-world datasets, 14 of which are firstly used in this paper and are made publicly available for future benchmarks.
The experimental results reveal that, affected by the discrepancy problem, the direct adaptations of the corresponding MLC methods fail to obtain better accuracy than the baseline approach that performs independent regressions. 
On the other hand, the extended versions obtain consistent improvements against the baseline, confirming the effectiveness of the proposed solution. 
Furthermore, extended versions of ERC obtain significantly better accuracy than state-of-the-art methods, including a method based on ensembles of multi-objective decision trees \citep{kocev2007} and a recent regularization-based multi-task learning method \citep{jalali+etal:2010,jalali+etal:2013}. 
Moreover, it is shown that, compared to the rest of the methods, the extended versions of ERC are associated with the smallest risk of decreasing the accuracy of the baseline, an appealing property.

The rest of the paper is organized as follows: Section~\ref{sec:methods} presents the SST and ERC methods and describes the discrepancy problem and the proposed solution. 
Section~\ref{sec:related} discusses related work from the MTR field, including well-known statistical procedures and multi-task learning methods, and points out differences with previous work on the discrepancy problem.
The details of the experimental setup (method configuration, evaluation methodology, datasets) are given in Section~\ref{sec:setup} and Section~\ref{sec:results} presents and discusses the experimental results. Finally, Section~\ref{sec:conclusions} offers our conclusion and outlines future work directions.
\section{Methods}
\label{sec:methods}

We first formally describe the MTR task and provide the notation that will be used subsequently for the description of the methods. Let $\vec{X}$ and $\vec{Y}$ be two random vectors where $\vec{X}$ consists of $d$ input variables $X_1,..,X_d$ and $\vec{Y}$ consists of $m$ target variables $Y_1,..,Y_m$. We assume that samples of the form $(\vec{x,y})$ are generated i.i.d. by some source according to a joint probability distribution $\vec{P}(\vec{X,Y})$ on $\mathcal{X} \times \mathcal{Y}$ where $\mathcal{X}=R^d$\footnote{$\mathcal{X}=R^d$ is used only for the sake of brevity. The domain of the input variables can also be discrete.} and $\mathcal{Y}=R^m$ are the domains of $\vec{X}$ and $\vec{Y}$ and are often referred to as the input and the output space. In a sample $(\vec{x,y})$, $\vec{x}=[x_1,..,x_d]$ is the input vector and $\vec{y}=[y_1,..,y_m]$ is the output vector which are realizations of $\vec{X}$ and $\vec{Y}$ respectively.
Given a set $D=\{(\vec{x}^1,\vec{y}^1),..,(\vec{x}^n,\vec{y}^n)\}$ of $n$ training examples, the goal in MTR is to learn a model $\vec{h}:\mathcal{X}\rightarrow\mathcal{Y}$ that given an input vector $\vec{x}$, is able to predict an output vector $\vec{\hat{y}} = \vec{h(x)}$ that best approximates the true output vector $\vec{y}$.

In the baseline Single-Target (ST) method, a multi-target model $\vec{h}$ is comprised of $m$ single-target models $h_j:\mathcal{X} \rightarrow R$ where each model $h_j$ is trained on a transformed training set $D_j = \{(\vec{x}^{1},y_{j}^{1}),..,(\vec{x}^{n},y_{j}^{n})\}$ to predict the value of a single target variable $Y_j$. This way, target variables are modeled independently and no attempt is made to exploit potential dependencies between them.
Despite the simplicity of the ST approach, several empirical studies 
\citep[e.g.][]{luaces2012binary} have shown that Binary Relevance, its MLC counterpart, often obtains comparable performance with more sophisticated MLC methods that model label dependencies, especially in cases where the underlying single-target prediction model is well fitted to the data \citep{dembczynski2012,read2014deep,read2015multi}. A theoretical explanation of these results was offered by \cite{dembczynski2012} who showed that modeling the marginal conditional distributions $P(Y_i|\mathbf{x})$ of the labels (as done by Binary Relevance) can be sufficient for getting good results in multi-label losses whose risk minimizers can be expressed in terms of marginal distributions (e.g. Hamming loss).

\subsection{Stacked Single-Target}
\label{subsec:mtrs}

Stacked Single-Target (SST) is inspired from the Stacked Binary Relevance method \citep{godbole2004} where the idea of stacked generalization \citep{wolpert1992stacked} was applied in a MLC context.
The training of SST consists of two stages. In the first stage, $m$ independent single-target models $h_j:\mathcal{X} \rightarrow R$ are learned as in ST. However, instead of directly using these models for prediction, SST involves an additional training stage where a second set of $m$ meta models $h'_j: \mathcal{X} \times R^{m} \rightarrow R$ are learned, one for each target $Y_j$.
Each meta model $h'_j$ is learned on a transformed training set $D'_j=\{(\vec{x}'^{1},y^1_j),\dots,(\vec{x}'^{n},y^n_j)\}$, where the original input vectors of the training examples ($\vec{x}^{i}$) have been augmented by estimates of the values of their target variables ($\hat{y}^i_1,\ldots,\hat{y}^i_m$) to form expanded input vectors $\vec{x}'^{i}=[\vec{x}^{i},\hat{y}^i_1,\ldots,\hat{y}^i_m]$. These estimates are obtained by applying the first stage models to the examples of the training set.

\begin{figure}
\centering
\fbox{
\resizebox{0.96\textwidth}{!}{
\includegraphics[trim = 0cm 2cm 8cm 0cm, clip]{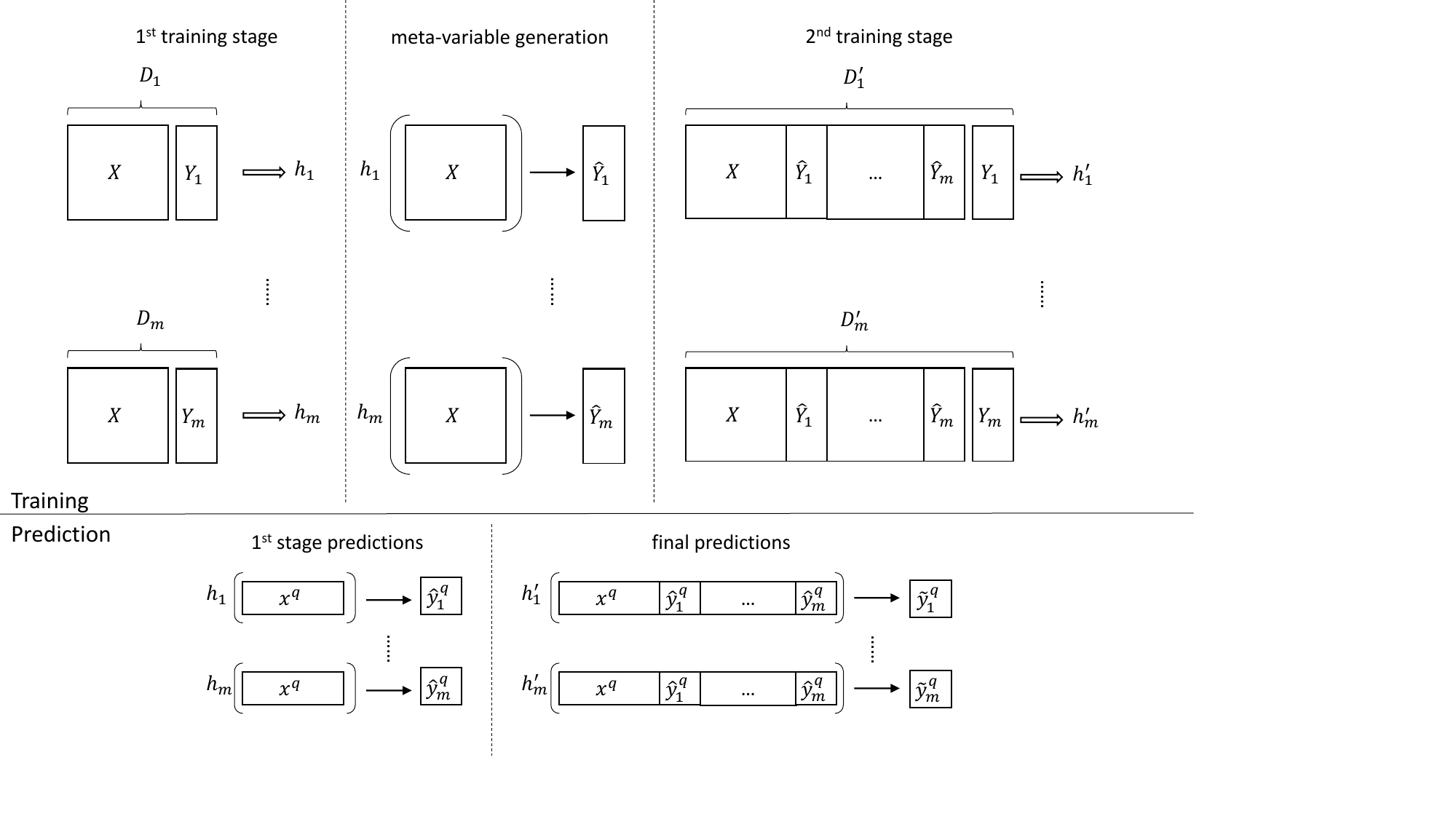}
}
}
\caption{Graphical illustration of SST's training and prediction procedures.}
\label{fig:sst_diagram}
\end{figure}

To obtain predictions for an unknown instance $\vec{x}^q$, the first stage models are first applied and an output vector $\hat{\vec{y}}^q=[h_1(\vec{x}^q),..,h_m(\vec{x}^q)]$ is obtained. Then, the second stage models are applied on transformed input vectors $\vec{x}'^{q} =[\vec{x}^q,\hat{\vec{y}}^q]$ to produce the final output vector $\tilde{\vec{y}}^q = [h'_1(\vec{x}'^{q}_1),\ldots,h'_m(\vec{x}'^{q}_m)]$. 
The training and prediction procedures of SST are graphically illustrated in Figure~\ref{fig:sst_diagram}.

\subsection{Ensemble of Regressor Chains}
\label{subsec:erc}

Regressor Chains (RC) is derived from Classifier Chains \citep{read2011}, a recently proposed MLC method based on the idea of chaining binary models. The training of RC consists of selecting a random chain (permutation) of the set of target variables and then building a separate regression model for each target. Assuming that the chain $C = \{Y_1 , Y_2 , .. , Y_m\}$ ($C$ represents an ordered set) is selected, the first model concerns the prediction of $Y_1$, has the form $h_1: \mathcal{X} \rightarrow R$ and is the same as the model built by the ST method for this target. The difference in RC is that subsequent models $h_{j}, j>1$ are learned on transformed training sets $D'_j = \{(\vec{x}'^{1}_j,y^{1}_{j}),..,(\vec{x}'^{n}_j,y^{n}_{j})\}$, where the original input vectors of the training examples have been augmented by the actual values of all previous targets of the chain to form expanded input vectors $\vec{x}'^{i}_j = [x^i_1,..,x^i_d,y^{i}_{1},..,y^{i}_{j-1}]$. Thus, the models built for targets $Y_{j}$ have the form $h_j: \mathcal{X} \times R^{j-1} \rightarrow R$. 

\begin{figure}
\centering
\fbox{
\resizebox{0.96\textwidth}{!}{
\includegraphics[trim = 1.5cm 5cm 9cm 1.5cm, clip]{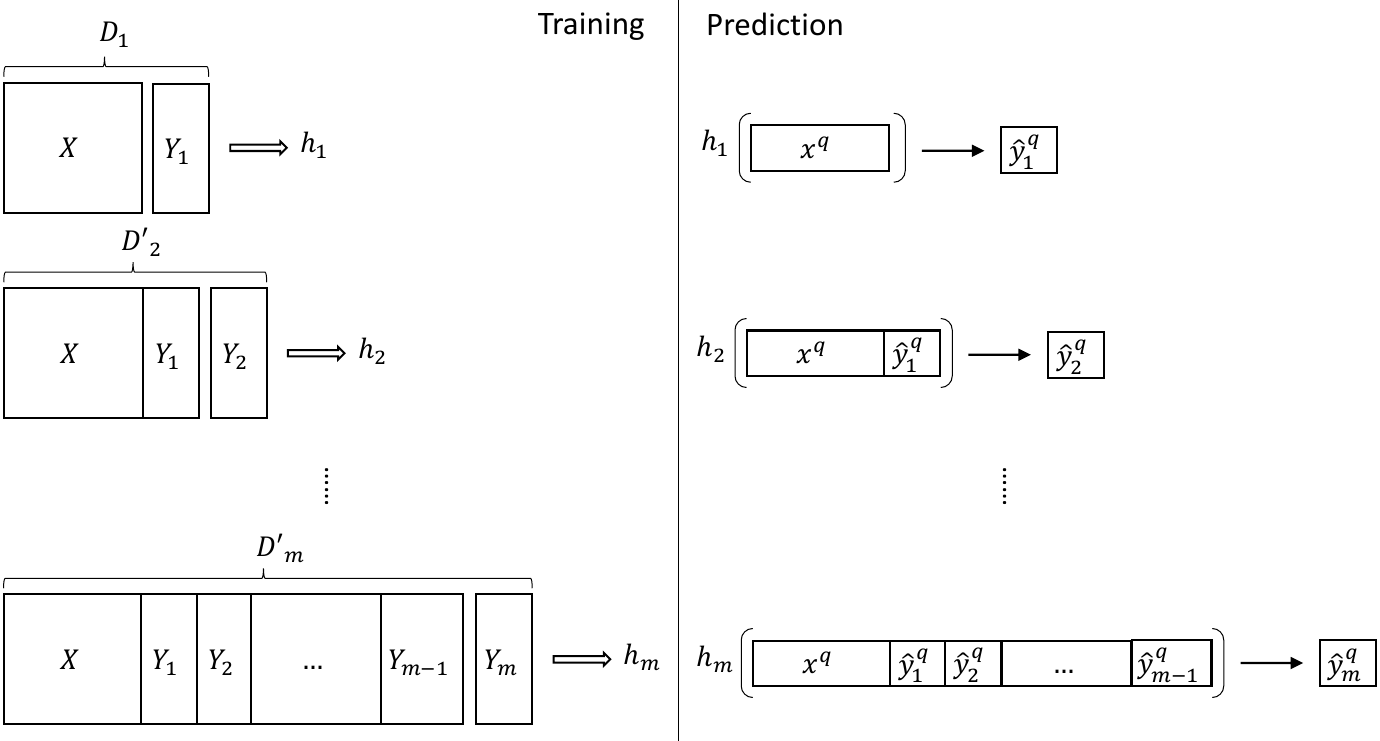}
}
}
\caption{Graphical illustration of RC's training and prediction procedures.}
\label{fig:rc_diagram}
\end{figure}

Given such a chain of models, the output vector $\hat{\vec{y}}^q$ of an unknown instance $\vec{x}^q$ is obtained by sequentially applying the models $h_j$, thus $\hat{\vec{y}}^q = [h_1(\vec{x}^q),h_2(\vec{x}'^{q}_2),..,h_m(\vec{x}'^{q}_m)]$ where $\vec{x}'^{q}_{j}= [x^q_1,..,x^q_d,\hat{y}^q_1,..,\hat{y}^q_{j-1}]$. Note that since the true values $y^q_1,..,y^q_{j-1}$ of the target variables are not available at prediction time, the method relies on estimates of these values obtained by applying the models $h_1,..,h_{j-1}$.
The training and prediction procedures of RC are graphically illustrated in Figure~\ref{fig:rc_diagram}.

One notable property of RC is that it is sensitive in the selected chain ordering. 
To alleviate this issue, \cite{read2011} proposed an ensemble scheme called Ensemble of Classifier Chains where a set of $k$ Classifier Chains models with different random chains are built on bootstrap samples of the training set and the final predictions come from majority voting. This scheme has been shown to consistently improve the accuracy of a single Classifier Chain in the classification domain. We apply the same idea on RC and compute the final predictions by taking the mean of the $k$ estimates for each target. The resulting method is called \textit{Ensemble of Regressor Chains} (ERC).

\subsection{Theoretical Insights into Stacking and Chaining}
\label{sec:methods:theory}

Both stacking and chaining have enjoyed significant attention from the MLC community, mainly due to their high performance and conceptual simplicity. A number of recent works have attempted a theoretical analysis of the methods \citep{dembczynski2010a,dembczynski2012,read2015multi}.
Adopting a statistical perspective, the authors of \citep{dembczynski2010a,dembczynski2012} distinguish between two types of label dependence:
\begin{itemize}
\item[-]unconditional, where $P(\mathbf{Y}) \neq \prod_{i=1}^{m} P(Y_i)$; and 
\item[-]conditional, where $P(\mathbf{Y}|\mathbf{x}) \neq \prod_{i=1}^{m} P(Y_i|\mathbf{x})$,
\end{itemize}
and show that modeling them is important for improving generalization performance. 
According to this analysis, stacking is interpreted as a method that models unconditional label dependence and is more suitable for minimizing label-wise decomposable multi-label loss functions\footnote{Note, however, that this analysis concerns a version of stacking that does not include the original input variables in the input space of the second stage models.}, while chaining is interpreted as a method that models conditional dependence and is more suitable for minimizing multi-label loss functions that cannot be decomposed label-wise.

Another interesting interpretation is offered by \cite{read2015multi} who show that Binary Relevance can (under certain conditions) achieve optimal performance in any dataset, and that improvements over the independent approach are often the result of using an inadequate base learner. Under this view, stacking and chaining can be considered as `deep' independent learners who owe their improved performance over Binary Relevance (when the same base learner is used) to the use of
labels as nodes in the inner layers of a deep neural network. These nodes represent readily available\footnote{This is in contrast with traditional deep learning where high-level feature representations are typically learned from the data in an unsupervised way.} (in the training phase), high-level transformations of the original inputs.
This interpretation of stacking and chaining applies directly to the MTR versions of these methods that we present here.

From a bias-variance perspective, we observe that by introducing additional features to single-target models, SST and ERC have the effect of decreasing their bias at the expense of an increased variance. This suggests that whenever the increase in variance is outweighed by the decrease in bias, one should expect gains in generalization performance over ST. 
This also hints that both methods will probably benefit from being combined with a base regressor that includes a variance reduction mechanism like bagged \citep{breiman1996} regression trees\footnote{An explicit feature selection could alternatively be applied as a means of variance reduction.}. 
As shown in \citep{munson2009feature}, bagged trees not only ignore irrelevant features but can also exploit features that contain useful but noisy information.
Both properties are very important in the context of SST and ERC because some of the extra features that they introduce might be irrelevant (e.g. whenever two target variables are statistically independent) and/or noisy (as discussed in the following subsection).

\subsection{Generation of Meta-inputs}
\label{sec:methods:meta}

Both SST and ERC are based on the same core idea of treating other prediction targets as additional input variables. 
These meta-inputs differ from ordinary inputs in the sense that while their actual values are available at training time, they are missing during prediction. Thus, during prediction both methods have to rely on estimates of these values which come either from ST (in the case of SST) or from RC (in the case of ERC) models built on the training set. An important question that is answered differently by each method is the following: \textit{What type of values should be used at training time for the meta-inputs?} SST uses estimates of the variables obtained by applying the first stage models on the training examples, while ERC uses their actual values. We observe that in both cases a core assumption of supervised learning is violated: that the training and testing data should be identically and independently distributed.
In the SST case, the in-sample estimates that are used to form the training examples of the second stage models will typically be more accurate than the out-of-sample estimates used at prediction time. 
The situation is even more problematic in the case of ERC since the actual target values are used during training.
In both cases, some of the input variables that are used by the underlying regression algorithm during model induction, become noisy (or noisier in the case of SST) at prediction time and, as a result, the induced model might wrongly estimate (overestimate) their usefulness.

To mitigate this problem, we propose the use of out-of-sample estimates of the targets during training in order to increase the compatibility between the training values of the target variables and the values used during prediction.
One way to obtain such estimates is to use a subset of the training set for building the first stage ST models (in the case of SST) or the RC models (in the case of ERC) and apply them to the held-out part. However, this approach would lead to reduced second stage training sets for SST as only the examples of the held-out set would be available for training the second stage models. The same holds for ERC where the chained RC models would be trained on training sets of decreasing size.
The solution that we propose to this problem is the use of an internal $f$-fold cross-validation approach that allows obtaining out-of-sample estimates of the target variables for all the training examples.
Compared to the actual target values or the in-sample estimates of the targets, the cross-validation estimates are expected to better resemble the values that are used during prediction. As a result, we expect that the contribution of the meta-inputs to the prediction of each target will be better estimated by the underlying regression algorithm.

The training procedures of the extended SST (denoted as SST$_{cv}$) and RC (denoted as RC$_{cv}$) methods are outlined in Algorithms~\ref{algorithm:mts:training} and \ref{algorithm:rc:training}. ERC$_{cv}$ consists of simply repeating the RC$_{cv}$ procedure $k$ times with different random chains. 
The corresponding prediction procedures are presented in Algorithms~\ref{algorithm:mts:prediction} and \ref{algorithm:rc:prediction}.  Note that the prediction procedures of the original and the extended versions of each method coincide.
In Section~\ref{sec:results} we compare the performance of the extended versions
of SST and ERC with the performance of the directly adapted variants, henceforth denoted as SST$_{train}$ and ERC$_{true}$. To better study the effects of the discrepancy problem, the comparison also includes SST using the actual target values (SST$_{true}$) and ERC using in-sample estimates of the target variables (ERC$_{train}$).

\begin{algorithm}
\caption{$SST_{cv}$ training}
\label{algorithm:mts:training}
\DontPrintSemicolon
    \KwIn{Training set $D$, number of internal cross-validation folds $f$}
    \KwOut{1st \& 2nd stage models $h_j$ \& $h'_j$, $j=1..m$}
    \BlankLine

    \tcp*[l]{\emph{Build 1st stage models}}
    \For{$j=1$ \KwTo $m$}
    {
        $D_{j} = \{(\vec{x}^{1},y^{1}_{j}),..,(\vec{x}^{n},y^{n}_{j})\}$
        \tcp*[r]{transform $D$ to $D_{j}$}

        $h_j: D_{j} \rightarrow R$
        \tcp*[r]{build model for $Y_j$ using $D_{j}$}

        split $D_j$ randomly into $f$ disjoint parts $D^i_j, i=1..f$ \;


        \For{$i = 1$ \KwTo $f$}
        {
            $h^i_j: D_j \setminus D^i_j \rightarrow R$
            \tcp*[r]{build model for $Y_j$ using $D_j \setminus D^i_j$}
        }
        
    }

    \tcp*[l]{\emph{Generate 2nd stage training sets}}
    \For{$j=1$ \KwTo $m$}{
		$D'_j \gets \emptyset$
           
        \For{$i=1$ \KwTo $f$}{
			$D'^{i}_j \gets \emptyset$
        
            \ForEach{$\vec{x}^k \in D^i_j$}{
				$\hat{\vec{y}}^k = [h^i_1(\vec{x}^k),..,h^i_m(\vec{x}^q)]$ \;           
            	$\vec{x}'^k = [\vec{x}^k,\hat{\vec{y}}^{k}]$
            	\tcp*[r]{concatenate $\vec{x}^k$ and $\hat{\vec{y}}^{k}$}
  
                $D'^{i}_j = D'^{i}_j \cup (\vec{x}'^{k}, y^k_j)$ \;
            }
            
            $D'_j = D'_j \cup D'^{i}_j$ \;
        }
    }

    \tcp*[l]{\emph{Build 2nd stage models}}
    \For{$j=1$ \KwTo $m$}{
         $h'_j: D'_{j} \rightarrow R$ \;
    }
\end{algorithm}

\begin{algorithm}
\caption{$SST$ prediction}
\label{algorithm:mts:prediction}
\DontPrintSemicolon
    \KwIn{Unknown instance $x^q$, 1st \& 2nd stage models $h_j$ \& $h'_j$, $j=1..m$}
    \KwOut{Output vector $\tilde{\vec{y}}^q$}
    \BlankLine

    $\hat{\vec{y}}^q = \tilde{\vec{y}}^q  = \vec{0}$

    \tcp*[l]{\emph{Apply the 1st stage models}}
    \For{$j = 1$ \KwTo $m$}{
         $\hat{y}^q_j=h_j(\vec{x}^q)$ \;
    }

    $\vec{x}'^{q} = [\vec{x}^q,\hat{\vec{y}}^q]$
    \tcp*[r]{concatenate $\vec{x}^q$ and $\hat{\vec{y}}^q$}

    \tcp*[l]{\emph{Apply the 2nd stage models}}
    \For{$j = 1$ \KwTo $m$}{
         $\tilde{\vec{y}}^q_j = h'_j(\vec{x}'^{q})$  \;
    }
\end{algorithm}
\begin{algorithm}
\caption{$RC_{cv}$ training}
\label{algorithm:rc:training}
\DontPrintSemicolon
    \KwIn{Training set $D$, number of internal cross-validation folds $f$}
    \KwOut{Chained models $h_j,j=1..m$}
    \BlankLine

    \tcp*[l]{\emph{Generate  $D'_{1}$}}
    $D'_{1} = \{(\vec{x}^{1},y^{1}_{1}),..,(\vec{x}^{n},y^{n}_{1})\}$
    \tcp*[r]{transform $D$ to  $D'_{1}$}

    \For{$j = 1$ \KwTo $m$}{

        $h_j: D'_{j} \rightarrow R$
        \tcp*[r]{build model for $Y_j$ using $D'_{j}$}

        \If{$j<m$}{

            \tcp*[l]{\emph{Generate  $D'_{j+1}$}}
            $D'_{j+1} \gets \emptyset$

            split $D'_j$ randomly into $f$ disjoint parts $D'^{i}_j, i=1..f$  \;

            \For{$i = 1$ \KwTo $f$}{

                $h^i_j: D'_{j} \setminus D'^{i}_{j} \rightarrow R$
                \tcp*[r]{build model for $Y_j$ using  $D'_{j} \setminus D'^{i}_{j}$}

                \ForEach{$\vec{x}'^{k}_j \in D'^{i}_{j}$}{
                    $x'^{i}_{j+1} = x'^{i}_j$

                    $\hat{y}^k_j= h^i_j(\vec{x}'^{k}_j)$ \;

                    $\vec{x}'^{k}_{j+1} = [\vec{x}'^{k}_j,\hat{y}^k_j]$
                    \tcp*[r]{append $\vec{x}'^{k}_j$ with  $\hat{y}^k_j$}

                    $D'_{j+1} = D'_{j+1} \cup (\vec{x}'^{k}_{j+1},y^k_{j+1})$ \;
                }
            }
        }
    }
\end{algorithm}

\begin{algorithm}
\caption{$RC$ prediction}
\label{algorithm:rc:prediction}
\DontPrintSemicolon
    \KwIn{Unknown instance $x^q$, chain models $h_j$, $j=1..m$}
    \KwOut{Output vector $\hat{\vec{y}}^q$}
    \BlankLine

    $\hat{\vec{y}}^q = \textbf{0}$

    $\vec{x}'^{q}_1 = \vec{x}^q$ \;
    \For{$j = 1..m$}{
        $\hat{y}^q_j=h_j(\vec{x}'^{q}_j)$ \;
        \If{$j<m$}{
        	$\vec{x}'^{q}_{j+1} = [\vec{x}'^{q}_j,\hat{y}^q_j]$
        	\tcp*[r]{append $\vec{x}'^{q}_j$ with $\hat{y}^q_j$}
		}    
    }

\end{algorithm}

\subsection{Discussion}
\label{sec:methods:discussion}

Besides the type of values that each method uses for the meta-inputs at training time, SST and ERC have additional conceptual differences.
A notable one is that the model built for each target $Y_j$ by SST, uses all other targets as inputs while in RC each model involves only targets that precede $Y_j$ in a random chain. As a result, the model built for $Y_j$ by RC, cannot benefit from statistical relationships with targets that appear later than $Y_j$ in the chain. This potential disadvantage of RC is partially overcome by ERC since each target is included in multiple random chains and, therefore, the probability that other targets will precede it is increased. 
At a first glance, SST seems to represent a more straightforward way of including all the available information about other targets. However, we should take into account that, since both methods rely on estimates of the meta-inputs at prediction time (as discussed in previous subsection), the more the meta-inputs that are included in the input space, the higher the amount of error accumulation that is risked at prediction time. From this perspective, ERC seems to adopt a more cautious approach than SST.
On the other hand, the estimates of the meta-inputs that are used by the second stage models in SST come from independent models, while the estimates of the meta-inputs used by each model in RC (and ERC) come from models that include information about other targets and thus involve a higher risk of becoming noisy.
Overall, there seems to be a trade-off between using the additional information available in the targets and the noise that this information comes with. Which of the two methods (and which variant) achieves a better balance in this trade-off is revealed by the experimental analysis in Section~\ref{sec:results}.

\subsection{Complexity Analysis}
\label{sec:methods:complexity}

In this section we discuss the time complexity of all variants of the proposed methods at training and test time, given a single-target regression algorithm with training complexity $O(g_{tr}(n,d))$ and test complexity $O(g_{te}(n,d))$ for a dataset with $n$ examples and $d$ input variables. The training and test complexities of the ST method are $O(m {\cdot} g_{tr}(n,d))$ and $O(m {\cdot} g_{te}(n,d))$ respectively, as it involves training and querying $m$ independent single-target models.

With respect to SST, the method builds $2 {\cdot} m$ models at training time, all of which are queried at prediction time. In all variants of the method, half of the models are built on the original input space and half of the models are built on an input space augmented by $m$ meta-inputs. Thus, in the case of SST$_{true}$, where the meta-inputs are readily available, the training and test complexities are $O(m {\cdot} (g_{tr}(n,d) {+} g_{tr}(n,d{+}m)))$ and $O(m {\cdot} (g_{te}(n,d) {+} g_{te}(n,d{+}m)))$ respectively. Given that in most cases (see Table~\ref{tbl:data sets}) the number of targets is much smaller than the number of inputs, i.e. $m \ll d$, the effective training and test complexities of SST$_{true}$ become $O(m {\cdot} g_{tr}(n,d))$ and $O(m {\cdot} g_{te}(n,d))$ respectively, thus same with ST's complexities. SST$_{train}$ and SST$_{cv}$ have the same test complexity with SST$_{true}$ but a larger training complexity because of the process of generating estimates for the meta-inputs. In the SST$_{train}$ case, the training complexity is $O(m {\cdot} g_{tr}(n,d) {+} m {\cdot} g_{te}(n,d))$ because the $m$ first-stage models are applied to obtain estimates for all the training examples. For most regression algorithms (e.g. regression trees), the computational cost of making predictions for $n$ instances is much smaller than the cost of training on $n$ examples. For instance, the training complexity of a typical binary regression tree learner is $O(n {\cdot} d^2)$ \citep{su2006fast} while the test complexity is $O(n {\cdot} \log_2 d)$. Thus, practically, the training complexity of SST$_{train}$ is similar to that of SST$_{true}$. When it comes to SST$_{cv}$, in addition to the $m$ first-stage models, $f$ additional models are built on $\frac{f-1}{f} {\cdot} n$ examples each. Therefore, the training complexity of SST$_{cv}$ is  $O(m {\cdot} g_{tr}(n,d) + m {\cdot} f {\cdot} g_{tr}(\frac{f-1}{f} {\cdot} n,d) + m {\cdot} g_{te}(n,d)) \approx O(f {\cdot} m {\cdot} g_{tr}(n,d) {+} m {\cdot} g_{te}(n,d))$. Given
that $g_{te}(n,d)$ $\ll$ $g_{tr}(n,d)$, we conclude that the training complexity of SST$_{cv}$ is roughly $f$ times ST's training complexity. 
Also, note that SST$_{train}$ and SST$_{cv}$ can be parellelized stage-wise both at training and at prediction time, i.e. all single-target models within the same level can be trained and queried independently, while SST$_{true}$ is fully parallelizable at training time (all single-target models can be trained independently) and stage-wise parallelizable at test time.

In ERC, each RC model consists of a chain of $m$ models built on input spaces augmented by $\{0,1,\ldots,m-1\}$ meta-inputs, thus $\frac{m-1}{2}$ meta-inputs on average. In the case of ERC$_{true}$ and for an ensemble size of $k$ RC models, the training and test complexities are $O(k {\cdot} m {\cdot} g_{tr}(n,d{+}(\frac{m-1}{2}))$ and $O(k {\cdot} m {\cdot} g_{te}(n,d{+}(\frac{m-1}{2}))$ respectively.  
Given, as before, that $m \ll d$, the training complexity of ERC$_{true}$ becomes $O(k {\cdot} m {\cdot} g_{tr}(n,d))$ and its test complexity becomes $O(k {\cdot} m {\cdot} g_{te}(n,d))$, thus $k$ times ST's complexity in both cases. Following a similar reasoning as we did above for SST, we can show that the training complexity of ERC$_{train}$ is similar to that of ERC$_{true}$ and that the training complexity of ERC$_{cv}$ is $O(k {\cdot} f {\cdot} m {\cdot} g_{tr}(n,d))$, i.e. $k {\cdot} f $ times ST's training complexity. Obviously, the test complexities of both ERC$_{train}$ and ERC$_{cv}$ are the same as ERC$_{true}$'s test complexity. With respect to parellelization, we observe that each member of an ERC$_{train}$ or ERC$_{cv}$ ensemble can be trained independently, while ERC$_{true}$ is fully parallelizable at training time, i.e. all $k {\cdot} m$ single-target models can be trained independently. For all ERC variants, test time parallelization is also possible since each ensemble member can be queried independently. 

\begin{table}[t!]
\centering
\caption{{\small Training and test complexities of the proposed methods with single- and multi-core implementations. $n$, $d$ and $m$ denote the numbers of data points, inputs, and targets respectively. $k$ denotes the number of chains in ERC and $f$ the number of internal cross-validation folds in the $cv$ variants of SST and ERC.}}
\begin{tabular}{cc|cc|cc}
\hline
\multicolumn{2}{c|}{\multirow{2}{*}{Method}} & \multicolumn{2}{c|}{Training complexity} & \multicolumn{2}{c}{Test complexity} \\
\cline{3-6}
& & \multicolumn{1}{c}{single-core} & \multicolumn{1}{c|}{multi-core} & \multicolumn{1}{c}{single-core} & \multicolumn{1}{c}{multi-core} \\
\hline
\multirow{3}{*}{\rotatebox{90}{SST}} & $true$ & $O(m {\cdot} g_{tr}(n,d))$ & $O(g_{tr}(n,d))$ & $O(m {\cdot} g_{te}(n,d))$ & $O(g_{te}(n,d))$ \\
& $train$ &  $O(m {\cdot} g_{tr}(n,d))$ &  $O(g_{tr}(n,d))$ & $O(m {\cdot} g_{te}(n,d))$ & $O(g_{te}(n,d))$ \\
& $cv$ & $O(f {\cdot} m {\cdot} g_{tr}(n,d))$ &  $O(g_{tr}(n,d))$ & $O(m {\cdot} g_{te}(n,d))$ & $O(g_{te}(n,d))$ \\
\hline
\multirow{3}{*}{\rotatebox{90}{ERC}} & $true$ & $O(k {\cdot} m {\cdot} g_{tr}(n,d))$ & $O(g_{tr}(n,d))$ & $O(k {\cdot} m {\cdot} g_{te}(n,d))$ & $O(m {\cdot} g_{te}(n,d))$ \\
& $train$ & $O(k {\cdot} m {\cdot} g_{tr}(n,d))$ & $O( m {\cdot} g_{tr}(n,d))$ & $O(k {\cdot} m {\cdot} g_{te}(n,d))$ & $O(m {\cdot} g_{te}(n,d))$ \\
& $cv$ & $O(k {\cdot} f {\cdot} m {\cdot} g_{tr}(n,d))$ & $O( m {\cdot} g_{tr}(n,d))$ & $O(k {\cdot} m {\cdot} g_{te}(n,d))$ & $O(m {\cdot} g_{te}(n,d))$ \\
\hline
\end{tabular}
\label{tbl:complexity}
\end{table}

Table~\ref{tbl:complexity} summarizes the training and test complexities of each method assuming a single-core implementation as well as the minimum possible complexity when a multi-core implementation is used. Note that, as shown in the table, SST$_{cv}$ and ERC$_{cv}$ have the same multi-core complexity with SST$_{train}$ and ERC$_{train}$ respectively because their internal cross-validation procedure can also be parallelized.
\section{Related Work}
\label{sec:related}

\subsection{Multi-Target Regression}
\label{sec:related:mtr}

MTR was first studied in Statistics under the term multivariate regression with Reduced Rank Regression (RRR) \citep{Izenman1975}, FICYREG \citep{van1980multivariate} and two-block PLS \citep{wold1985partial} (the multiple response version of PLS) being three of the earliest methods. Among these methods, two-block PLS has been used more widely, especially in Chemometrics.
More recently, the Curds and Whey (C\&W) method was proposed \citep{breiman1997} and was found to outperform  RRR, FICYREG and two-block PLS.
As noted by \cite{breiman1997}, C\&W, RRR and FICYREG can all be expressed using the same generic form $\mathbf{\tilde{y}= B \hat{y}}$, where $ \mathbf{\hat{y}}$ are estimates obtained by applying ordinary least squares regression on the target variables and $\mathbf{B}$ is a matrix that modifies these estimates in order to obtain a more accurate prediction $\mathbf{\tilde{y}}$, under the assumption that the targets are correlated. 
In all methods, $\mathbf{B}$ can be expressed as $\mathbf{B = \mathbf{\hat{T}}^{-1} \mathbf{D} \mathbf{\hat{T}}}$, where $\mathbf{\hat{T}}$ is the matrix of sample canonical co-ordinates  and $\mathbf{D}$ is a diagonal ``shrinking'' matrix that is obtained differently in each method.
SST is highly similar to these methods but allows a more general formulation of the MTR problem. Firstly, SST does not impose any restriction to the family of models that generate the uncorrected (first stage) estimates in contrast to these approaches that use estimates obtained from least squares regression. Secondly, the correction of the estimates applied by SST comes from a learning procedure that jointly considers target and input variables rather than target variables alone.

As shown by \cite{breiman1997}, the above methods can be described by an alternative but equivalent scheme. According to this, $\mathbf{y}$ is first transformed to the canonical co-ordinate system $\mathbf{y'} = \mathbf{\hat{T}} \mathbf{y}$, then separate least squares regression is performed on each $\mathbf{y'}$ to obtain $\mathbf{\hat{y}'}$, these estimates are scaled by $\mathbf{D}$ to obtain  $\mathbf{\tilde{y}'} = \mathbf{D} \mathbf{\hat{y}'}$ and finally transformed back to the original output space $\mathbf{\tilde{y}} = \mathbf{\hat{T}^{-1}} \mathbf{\tilde{y}'}$.
As discussed by \cite{dembczynski2012}, from this perspective, these methods fall under a more general scheme where the output space is first transformed, single-target regressors are then trained on the transformed output space and an inverse transformation is performed (possibly along with shrinkage/regularization) to obtain predictions for the original targets. Due to its generality, this scheme has been adopted by a number of recent methods in both MLC \citep{hsu+etal:2009,zhang+schneider:2011,zhang+schneider:2012,tai+lin:2012} and MTR \citep{balasubramanian2012,tsoumakas:2014:ecmlpkdd}. 

A large number of MTR methods are derived from the predictive clustering tree (PCT) framework \citep{blockeel1998}. 
The main difference between the PCT algorithm and a standard decision tree is that the variance and the prototype functions are treated as parameters that can be instantiated to fit the given learning task. Such an instantiation for MTR tasks are the multi-objective decision trees (MODTs) where the variance function is computed as the sum of the variances of the targets, and the prototype function is the vector mean of the target vectors of the training examples falling in each leaf \citep{blockeel1998,blockeel1999}. Bagging and random forest ensembles of MODTs were developed by \cite{kocev2007} and were found significantly more accurate than MODTs and equally good or better than ensembles of single-objective decision trees for both regression and classification tasks. In particular, multi-objective random forests yielded better performance than multi-objective bagging.

Methods that deal with the prediction of multiple target variables can be found in the literature of the related learning task of multi-task learning.
According to \cite{caruana1997multitask}, multi-task learning is a form of inductive transfer \citep{pratt1993discriminability} where the aim is to improve generalization accuracy on a set of related tasks by using a shared representation that exploits commonalities between them.
This definition implies that a multi-task method should be able to deal with problems where different prediction tasks do not necessarily share the same set of training examples or descriptive features and, moreover, each task can have a different data type. Thus, multi-task learning is actually a generalization of MTR.

Artificial neural networks (ANNs) are very well suited for multi-task problems because they can be naturally extended to support multiple outputs and offer flexibility in defining how inputs are shared between tasks. 
Thus, it is not surprising that most of the earliest multi-task methods were based on ANNs. \cite{caruana1995learning}, for example, proposed a method where backpropagation is used to train single ANN with multiple outputs (connected to the same hidden layers), and showed that it has better generalization performance compared to multiple single-task ANNs. A different architecture was used by \citep{baxter1995learning} where only the first hidden layers are shared and subsequent layers are specific to each task. The question of how much sharing is better when multi-task ANNs are applied for stock return prediction was explored by \citep{ghosn1997multi} who concluded that a partial sharing of network parameters is preferable compared to full or no sharing.
More recently, \cite{collobert2008unified} applied a deep multi-task neural network architecture for natural language processing.

A large number of multi-task learning methods stem from a regularization perspective\footnote{A nice categorization of regularization-based multi-task methods can be found in \citep{zhou_chen_ye_2012}.}. Regularization-based multi-task methods minimize a penalized empirical loss of the form $\displaystyle \min_W \mathcal{L}(W)+\Omega(W)$, where $W$ is a parameter matrix that has to be estimated, $\mathcal{L}(W)$ is an empirical loss calculated on the training data and $\Omega(W)$ is a regularization term that takes a different form in each method depending on the underlying task relatedness assumption. 
Most methods assume that all tasks are related to each other \citep{evgeniou2004regularized,ando2005framework,NIPS2006_3143,argyriou2008convex,chen2009convex,chen2010learning,obozinski2010joint}, while there are methods assuming that tasks are organized in structures such as clusters \citep{jacob2009clustered,zhou2011clustered}, trees \citep{kim2010tree} and graphs \citep{chen2010graph}.
A well-studied category of methods, which are particularly useful when dealing with high-dimensional input spaces, assume that models for different tasks share a common low-rank subspace and impose a trace-norm constraint on the parameter matrix \citep{NIPS2006_3143,argyriou2008convex,ji2009accelerated}.
A similar category of methods constraint all models to share a common set of features (thus performing a joint feature selection), typically by applying $L_1 / L_q$-norm ($q>1$) regularization \citep{obozinski2010joint}. An approach that relaxes the above restrictive constraint allowing models to leverage different extents of feature sharing is proposed in \citep{jalali+etal:2010,jalali+etal:2013}.

Finally, we would like to mention that a number of MTR methods are based on the Gaussian Processes framework \citep[e.g.,][]{bonilla2007multi,alvarez2011}. These methods capture correlations between tasks by appropriate choices of covariance functions. A nice review of such methods as well as their relations to regularization-based multi-task approaches can be found in \citep{alvarez2011kernels}.

\subsection{Discrepancy in Meta-inputs}
\label{sec:related:discrepancy}

In the MLC domain, \cite{senge2012} studied how the discrepancy issue affects the performance of Classifier Chains and showed that longer chains (i.e. multi-label problems with more labels to be predicted) lead to a higher performance deterioration. 
In an extension of that work \cite{senge2013}, a ``rectified'' version of Classifier Chains (called Nested Stacking) was presented that uses in-sample estimates of the label variables for training as in Stacked Binary Relevance. It was shown that this method performs better than the original Classifier Chains, especially when the label dependencies are strong.
Following the opposite direction, \cite{montanes2011} proposed AID, a method similar to Stacked Binary Relevance, and found that using the actual label values instead of (in-sample) estimates, leads to better results for most multi-label evaluation measures in both AID and Stacked Binary Relevance. 

Our work is the first to study this issue in the MTR domain\footnote{Actually, an early version of this work \cite{spyromitros:2012:arxiv} is the first to consider the discrepancy problem in the context of input space expansion methods.}. The issue is studied jointly for SST and ERC, thus allowing general conclusions to be drawn for this type of methods. 
Furthermore, \cite{montanes2011,senge2013} compared only the use of actual target values with the use of in-sample estimates while our comparison includes the use of out-of-sample estimates obtained by a cross-validation procedure. Finally, \cite{senge2013} evaluate the use of estimates in Classifier Chains whereas we focus on the ensemble version of the corresponding MTR method (ERC) that is expected to offer more resilience to error propagation, as discussed in Section~\ref{sec:methods:discussion}.
\section{Experimental Setup}
\label{sec:setup}

This section describes our experimental setup. We first present the participating methods and their parameters and provide details about their implementation in order to facilitate reproducibility of the experiments. Next, we describe the
evaluation measure and explain the process that was followed for the statistical comparison of the methods. Finally, we present the datasets that we used and their main statistics.

\subsection{Methods, Parameters and Implementation}
\label{sec:experiments:methods}

The experimental evaluation includes all variants of the proposed SST and ERC methods, the ST baseline and the following state-of-the-art multiple prediction methods: a) multi-objective random forest (MORF) \citep{kocev2007}, b) trace norm regularization for multi-task learning (TNR) \citep{argyriou2008convex}, c) the \textsc{Dirty} approach for multi-task learning \citep{jalali+etal:2010,jalali+etal:2013} and d) a very recent multi-target method based on random linear combinations of the output space (RLC) \citep{tsoumakas:2014:ecmlpkdd}.
For easy reference, Table~\ref{tbl:methods:mtr} lists all methods included in the evaluation along with their abbreviations and citations where appropriate.

\begin{table}
\centering
\caption{Methods used in experiments with abbreviations and citations.}
\begin{tabular}{llr}
\hline
Abbr. & Method & Citation \\
\hline
ST & Single Target & \\
SST$_{true}$ & Stacked ST, true values & This paper \\
SST$_{train}$ & Stacked ST, in-sample estimates  & This paper \\
SST$_{cv}$ & Stacked ST, cv estimates & This paper \\
ERC$_{true}$ & Ensemble of Regressor Chains, true values & This paper \\
ERC$_{train}$ & Ensemble of Regressor Chains, in-sample estimates  & This paper \\
ERC$_{cv}$ & Ensemble of Regressor Chains, cv estimates & This paper \\
MORF & Multi-Objective Random Forest & \cite{kocev2007} \\
TNR  & Trace Norm Regularization multi-task learning & \cite{argyriou2008convex} \\
\textsc{Dirty}  & A Dirty model for multi-task learning & \cite{jalali+etal:2010,jalali+etal:2013} \\
RLC  & Random Linear target Combinations & \cite{tsoumakas:2014:ecmlpkdd} \\
\hline
\end{tabular}
\label{tbl:methods:mtr}
\end{table}

The proposed methods as well as ST and RLC transform the mutli-target regression task into a series of single-target regression tasks which can be dealt with using any standard regression algorithm.
For most of the experiments, we use bagged regression trees as the base regressor. This choice was motivated in Subsection~\ref{sec:methods:theory} and is further discussed in Subsection~\ref{sec:results:base} where we present results using a variety of well-known linear and non-linear regression algorithms.
The ensemble size of all ERC variants is set to $k=10$ RC models, each one trained using a different random chain. In datasets with less than 10 distinct chains, we create exactly as many RC models as the number of distinct chains. Furthermore, since the base regressor involves bootstrap sampling, we do not perform sampling in ERC, i.e. each RC model is trained using all training examples.
In SST, we exclude the target being predicted by each second stage model from the input space of that model as we found that this choice improves slightly the performance of all variants of this method.
$f=10$ internal cross-validation folds are used in both SST$_{cv}$ and ERC$_{cv}$. 

Concerning the parameter settings of the competitive methods, 
in MORF we use an ensemble size of 100 trees and the values suggested by \cite{kocev2007} for the rest of its parameters.
In RLC, we generate $r=100$ new target variables by combining $k=2$ of the original target variables (after bringing them to the $[0,1]$ interval). As shown in \citep{tsoumakas:2014:ecmlpkdd}, these values lead to near optimal results.
In TNR, we minimize the squared loss function using the accelerated gradient method for trace norm minimization \citep{ji2009accelerated}. The regularization parameter is tuned by selecting among the values $\{10^r : r \in \{-3,...,3\}\}$ with internal 5-fold cross-validation. Before applying TNR, we apply z-score normalization and add a bias column as suggested in \citep{zhou2011malsar}.
Finally, \textsc{Dirty} is setup as suggested in \citep{jalali+etal:2013}: Input variables are scaled to the $[-1,1]$ range by dividing them with their maximum values. The regularization parameters $\lambda_b$ and $\lambda_s$ are tuned via internal 5-fold cross-validation (as in TNR).  As suggested in \citep{jalali+etal:2013}, we set $\lambda_b = c \sqrt{\frac{m \log d}{n}}$, where $c \in \{10^r : r \in \{-2,...,2\}\} $ is a constant. Each distinct value of $\lambda_b$ is paired with five values of $\lambda_s = \frac{\lambda_b}{1+ \frac{m-1}{4} i}, i \in \{0,1,2,3,4\}$, thus respecting the $\frac{\lambda_s}{\lambda_b} \in [\frac{1}{m},1]$ relationship dictated by the optimality conditions. In total, 25 different combinations of $\lambda_b$ and $\lambda_s$ are evaluated.

All the proposed methods and the evaluation framework were implemented in Java and integrated in Mulan\footnote{\url{http://mulan.sourceforge.net}} \citep{tsoumakas2011b} by expanding its functionality to multi-target regression. The implementation of all single-target regression algorithms that were used to instantiate problem transformation methods are taken from Weka\footnote{\url{http://www.cs.waikato.ac.nz/ml/weka}}.
With respect to the competing methods, RLC was already integrated in Mulan while for the purposes of this study we also integrated MORF (via a wrapper of the implementation offered in CLUS\footnote{\url{http://dtai.cs.kuleuven.be/clus/}}) as well as TNR and \textsc{Dirty} (via wrappers of the implementations offered in MALSAR \citep{zhou2011malsar}). Thus, all methods were evaluated under a common framework.
In support of open science, we created a github project\footnote{\url{https://github.com/lefman/mulan-extended}} that contains all our implementations, including code that facilitates easy replication of our experimental results.

\subsection{Evaluation}
\label{sec:experiments:evaluation}

The proposed methods aim at reducing the prediction error on every single target of a MTR problem. To measure the performance of a MTR model on each target variable we use Relative Root Mean Squared Error (RRMSE).
The RRMSE of a model $\vec{h}$ that has been induced from a train set $D_{train}$ is estimated based on a test set $D_{test}$ according to the following equation:
\begin{equation}
RRMSE(\vec{h},D_{test})
=
\sqrt
{
    \frac
        {
            \sum_{(\vec{x},\vec{y}) \in D_{test}} (\hat{y}_{j}-y_{j})^{2}
        }
        {
            \sum_{(\vec{x},\vec{y}) \in D_{test}} (\bar{Y}_{j}-y_{j})^{2}
        }
}
\end{equation}
where $\bar{Y}_{j}$ is the mean value of target variable $Y_{j}$ over $D_{train}$ and $\hat{y}_{j}$ is the estimation of the MTR model $\vec{h}$ for $Y_{j}$. More intuitively, RRMSE for a target is equal to the Root Mean Squared Error (RMSE) for that target divided by the RMSE of predicting the average value of that target in the training set.
RRMSE is estimated using $k$-fold cross-validation on all datasets, i.e. one RRMSE measurement is obtained on each fold and the final RRMSE is calculated as the average of those measurements. We use $k=10$ on all datasets, except those with more than 9000 examples where for computational reasons we use either $k=5$ (rf1 and rf2) or $k=2$ (scm1d and scm20d)\footnote{The reliability of the estimates obtained using $k=2$ and $k=5$ has been validated by checking the stability of the rankings of the methods when repeating the cross-validation experiment with different random seeds.}. 

To test the statistical significance of the observed differences between the methods, we follow the methodology suggested by \cite{demsar2006}.
To compare multiple methods on multiple datasets we use the Friedman test, the non-parametric alternative of the repeated-measures ANOVA. The Friedman test operates on the average ranks of the methods and checks the validity of the hypothesis (null-hypothesis) that all methods are equivalent. Here, we use an improved (less conservative) version of the test that uses the $F_f$ instead of the $\chi^2_F$ statistic  \citep{iman1980approximations}.
When the null-hypothesis of the Friedman test is rejected ($p < 0.01$), we proceed with the Nemenyi post-hoc test that compares all methods to each other in order to find which methods in particular differ from each other. Instead of reporting the outcomes of all pairwise comparisons, we employ the simple graphical presentation of the test's results introduced by \cite{demsar2006}, i.e. all methods being compared are placed in a horizontal axis according to their average ranks and groups of methods that are not significantly different (at a certain significance level) are connected (see Figure~\ref{fig:nemenyi:straightforward} for an example). To generate such a diagram, a critical difference (CD) should be calculated that corresponds to the minimum difference in average ranks required for two methods to be considered significantly different. CD for a given number of methods and datasets, depends on the desired significance level. Due to the known conservancy of the Nemenyi test \citep{demsar2006}, we use a 0.05 significance level for computing the CD throughout the paper.

As the above methodology requires a single performance measurement for each method on each dataset, it is not directly applicable to multi-target evaluation where we have multiple performance measurements (one for each target) for each method on each dataset.
One option is to take the average RRMSE (aRRMSE) across all target variables within a dataset as a single performance measurement. 
This choice, however, has the disadvantage that a very small or large error on a single target might dominate the average, thus obscuring performance differences on the target level.
Another option is to treat the RRMSE of each method on each target as a different performance measurement. In this case, Friedman test's assumption of independence between performance measurements might be violated.
In the absence of a better solution, we perform a two-dimensional analysis (as done e.g. by \cite{aho2012}) where statistical tests are conducted using both aRRMSE (\textit{per dataset analysis}) but also considering RRMSE per target as an independent performance measurement (\textit{per target analysis}).

\subsection{Datasets}
\label{sec:experiments:datasets}

Despite the numerous interesting applications of MTR, there are only few publicly available datasets of this kind - perhaps because most applications are industrial - and most experimental evaluations of MTR methods are based on a limited amount of datasets.  
For this study, much effort was made for the composition of a large and diverse collection of benchmark MTR datasets. In addition to 5 datasets that have been used in previous studies and are publicly available (edm, sf1, sf2, jura, wq), we also used 5 publicly available datasets (enb, slump, andro, osales, scpf) that have not been used for MTR benchmarking in the past. We also collected raw MTR data from a variety of interesting application domains and composed 8 new benchmark datasets (atp1d, atp7d, oes97, oes10, rf1, rf2, scm1d, scm20d). In total we collected 18 datasets and make them publicly available for future studies\footnote{\url{http://mulan.sourceforge.net/datasets-mtr.html}}. To the best of our knowledge, this is the largest collection of benchmark MTR datasets to date.

\begin{table}
\caption{Name, source, number of examples, number of input variables ($d$) and number of target variables ($m$) of the datasets used in the evaluation. The datasets marked with an asterisk are first used for MTR benchmarking in this paper to the best of our knowledge.}
\label{tbl:data sets}
\begin{center}
\begin{tabular}{llrrr}
\hline
Dataset & Source & Examples & $d$ & $m$ \\
\hline
edm	   & \cite{karalic1997}	 	 & 154		  & 16	  & 2  \\
sf1    & \cite{Lichman:2013}     & 323		  & 10    & 3  \\
sf2    & \cite{Lichman:2013}     & 1066		  & 10    & 3  \\
jura   & \cite{goovaerts1997}	 & 359		  & 15    & 3  \\
wq     & \cite{dzeroski2000}     & 1060		  & 16    & 14 \\
*enb   & \cite{tsanas2012accurate}& 768   	  & 8     & 2  \\
*slump & \cite{Yeh2007474}	       & 103   	  & 7     & 3  \\
*andro & \cite{hatzikos2008}	 & 49   	  & 30    & 6  \\
*osales & \cite{kaggle:onsales}	 & 639   	  & 413   & 12 \\
*scpf  	  & \cite{kaggle:scpf}	 & 1137   	  & 23    & 3  \\
*atp1d  & This paper	 & 337		  & 411   & 6  \\
*atp7d  & This paper	 & 296		  & 411   & 6  \\
*oes97  & This paper	 & 334		  & 263   & 16 \\
*oes10  & This paper	 & 403		  & 298   & 16 \\
*rf1    & This paper	 & 9125		  & 64    & 8  \\
*rf2    & This paper	 & 9125		  & 576   & 8  \\
*scm1d  & This paper	 & 9803		  & 280   & 16 \\
*scm20d & This paper	 & 8966		  & 61    & 16 \\
\hline
\end{tabular}
\end{center}
\end{table}

Table~\ref{tbl:data sets} reports the name (1st column), source (2nd column), number of examples (3rd column), number of input variables (4th column) and number of target variables (5th column) of each dataset. 
Detailed descriptions of all datasets are provided in Appendix~\ref{appendix:datasets}.
\section{Experimental Analysis} 
\label{sec:results}

In this Section we present an extensive experimental analysis of the performance of the proposed methods.
Subsection~\ref{sec:results:base} is devoted to an exploration of the performance of ST using various well-known regression algorithms.
The purpose of this investigation is to help us select an algorithm that works well on the studied datasets and use it as base regressor in all problem transformation methods (ST, SST, ERC and RLC) in subsequent experiments. At the same time, a challenging baseline performance level will be set for all multi-target methods. 
In Subsection~\ref{sec:results:straightforward} we evaluate SST$_{train}$ and ERC$_{true}$, the direct adaptations of the corresponding MLC methods, in order to see whether these variants obtain a competitive performance compared to ST and state-of-the-art multi-target methods.
Next, in Subsection~\ref{sec:results:variants} all three meta-input generation variants ($true$, $train$, $cv$) of SST and ERC are evaluated and compared to ST, shedding light into the impact of the discrepancy problem on each method. 
After the best performing variants of each method have been identified, Subsection~\ref{sec:results:soa} compares them with the state-of-the-art. The running times of all methods are reported and compared in Subsection~\ref{sec:results:runtimes}, and finally, this section ends with a discussion of the main outcomes of the experimental results (Subsection~\ref{sec:results:discussion}).

\subsection{Base Regressor Exploration}
\label{sec:results:base}

In this subsection we explore the performance of ST on the studied domains using a variety of regression algorithms. The goal of this exploration is to help us identify a regression algorithm that performs well across many domains, thus setting a challenging baseline performance level for the multi-target methods that we study next. The algorithm that will emerge as the best performer will be used to instantiate all problem transformation methods (ST, SST, ERC and RLC) in the rest of the experiments, facilitating a fair comparison between these methods.

We selected five well-known linear and non-linear regression algorithms to couple ST with, in particular we use: ridge regression \citep{hoerl1970} (\textsc{ridge}), regression tree \cite{breiman1984classification} (\textsc{tree}), L2-regularized support vector regression regression \citep{drucker1997} (\textsc{svr}), bagged \citep{breiman1996} regression trees (\textsc{bag}) and stochastic gradient boosting \citep{friedman1999} (\textsc{sgb}).
In \textsc{ridge} and \textsc{svr}, the regularization parameter was tuned (separately for each target) by applying internal 5-fold cross-validation and choosing the value that leads to the lowest root mean squared error among $\{10^r : r \in \{-4,...,2\}\}$.
In \textsc{bag} we combine the predictions of 100 \textsc{tree}s while in \textsc{sgb} we boost trees with four terminal nodes using a small shrinkage rate ($0.1$) and a large number of iterations ($100$), as suggested by \cite{hastie09}.

\begin{figure}
\centering
    \begin{subfigure}[b]{0.49\textwidth}
    	\resizebox{1\textwidth}{!}{
        \includegraphics[trim = 4.5cm 11.7cm 3.2cm 10cm, clip]{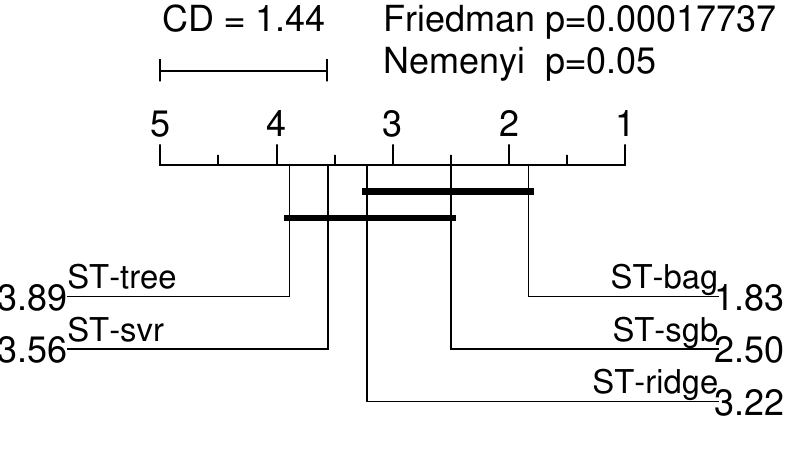}
        }
        \caption{Per dataset analysis.}
    \end{subfigure}
    ~
    \begin{subfigure}[b]{0.49\textwidth}
        \resizebox{1\textwidth}{!}{
        \includegraphics[trim = 4.5cm 11.7cm 3.2cm 10cm, clip]{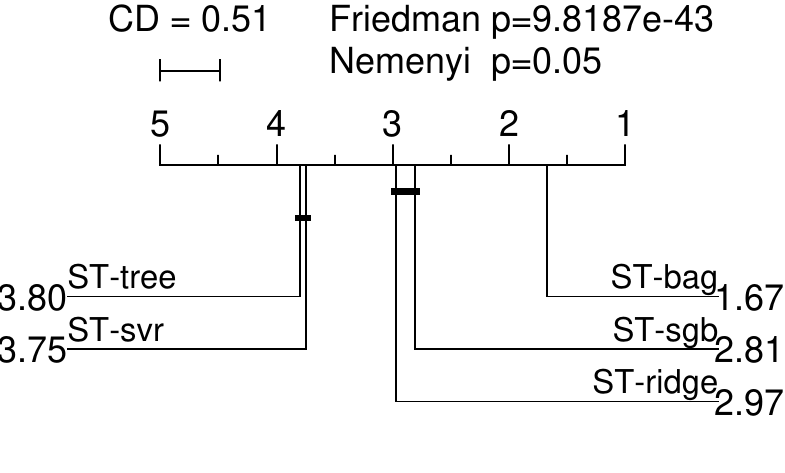}
        }
        \caption{Per target analysis.}     
    \end{subfigure}
    \caption{Comparison of different ST instantiations using the Nemenyi test. Groups of methods that are not significantly different (at $p=0.05$) are connected.}
	\label{fig:nemenyi:base}
\end{figure}

The detailed results obtained by each instantiation on each dataset and target are given in Appendix~\ref{appendix:results:st}. We observe that no algorithm is better in all domains (as dictated by the no free lunch theorems for supervised learning \citep{wolpert1996lack,wolpert2002supervised}). However, ST-\textsc{bag} stands out obtaining the lowest aRRMSE in nine datasets. ST-\textsc{sgb} follows with five wins while ST-\textsc{ridge} and ST-\textsc{svr} each obtain the lowest error in two datasets.
Figure~\ref{fig:nemenyi:base} shows the average ranks of the different instantiations along with the results of the Friedman and the Nemenyi tests for the analysis per dataset (left) and per target (right).
In both analyses, the lowest average rank is obtained by ST-\textsc{bag}, followed by ST-\textsc{sgb} and ST-\textsc{ridge}. In the per dataset analysis, the Nemenyi test finds that ST-\textsc{bag} is significantly better than ST-\textsc{tree} and ST-\textsc{svr} while in the per target analysis, ST-\textsc{bag} is found significantly better than all the other instantiations. 
Therefore, we use \textsc{bag} as the base regressor for all problem transformation methods in the rest of the experiments.

\subsection{Evaluation of Direct Adaptations}
\label{sec:results:straightforward}

In this subsection we focus on SST$_{train}$ and ERC$_{true}$, the versions of SST and ERC that use the same type of values for the meta-inputs as their MLC counterparts, and compare their performance to that of ST, MORF, RLC, TNR and \textsc{Dirty} to see where these methods stand with respect to the state-of-the-art.

\begin{figure}
\centering
    \begin{subfigure}[b]{0.49\textwidth}
    	\resizebox{1\textwidth}{!}{
        \includegraphics[trim = 4.5cm 10.8cm 3.2cm 10cm, clip]{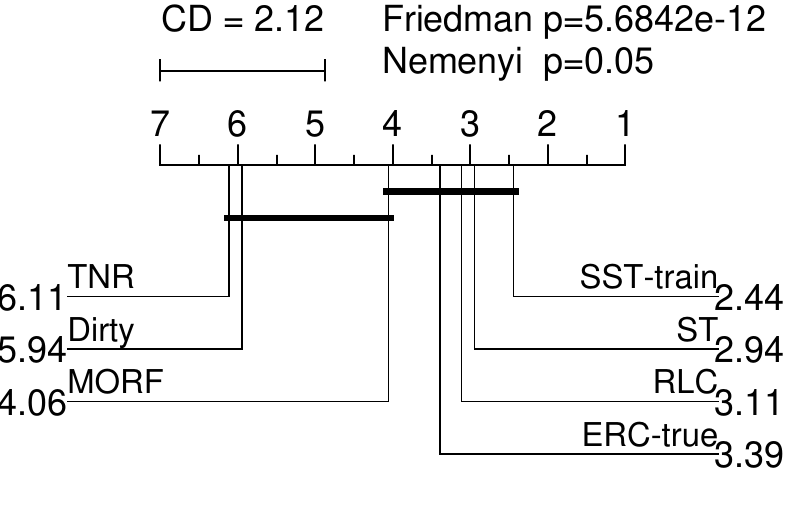}
        }
        \caption{Per dataset analysis.}
    \end{subfigure}
    ~
    \begin{subfigure}[b]{0.49\textwidth}
        \resizebox{1\textwidth}{!}{
        \includegraphics[trim = 4.5cm 10.8cm 3.2cm 10cm, clip]{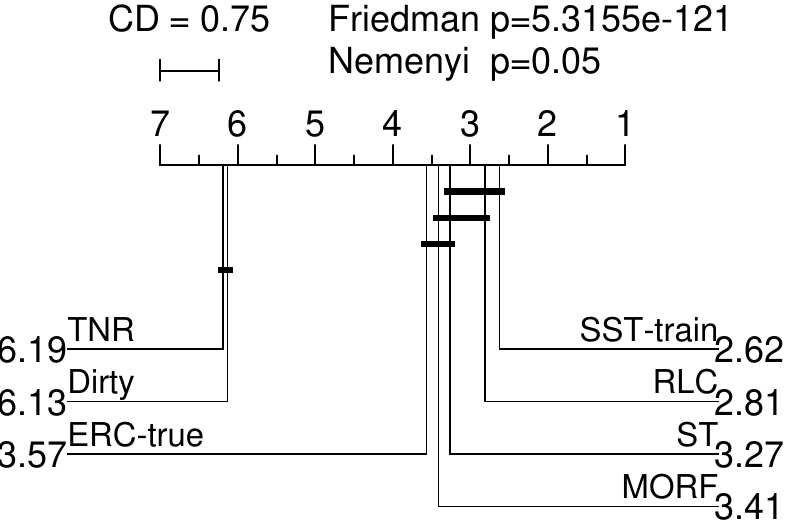}
        }
        \caption{Per target analysis.}   
    \end{subfigure}
    \caption{Comparison of direct adaptations (with \textsc{bag} as base regressor) using the Nemenyi test. Groups of methods that are not significantly different (at $p=0.05$) are connected.}
	\label{fig:nemenyi:straightforward}
\end{figure}

Figure~\ref{fig:nemenyi:straightforward} shows the average ranks of the methods along with the results of the Friedman and the Nemenyi tests when the analysis is performed per dataset (left) and per target (right)\footnote{The detailed results per dataset and target can be found in Appendix~\ref{appendix:results:mtr}}. Several interesting remarks can be made based on these results. 
First, we see that \textit{both SST$_{train}$ and ERC$_{true}$ are competitive with state-of-the-art methods}. SST$_{train}$ obtains the lowest average rank in both the per dataset and the per target analysis. In the per dataset analysis, it is found significantly better than TNR and \textsc{Dirty} and similar with MORF and RLC and in the per target analysis it is found better than TNR, \textsc{Dirty} and MORF and similar with RLC. ERC$_{true}$ performs worse than SST$_{train}$ but is still ranked above TNR, \textsc{Dirty} and MORF in the per dataset analysis and above TNR and \textsc{Dirty} in the per target analysis. In the per dataset analysis, ERC$_{true}$ is found significantly better than TNR and \textsc{Dirty} and similar with MORF and RLC, while in the per target analysis it is found better than TNR and \textsc{Dirty}, similar with MORF and only RLC outperforms it significantly.

Interestingly, however, we see that according to both the per dataset and the per target analysis, SST$_{train}$ and ERC$_{true}$ are not significantly better than ST. 
This is an indication that the use of targets as meta-input as implemented by these variants of SST and ERC does not bring significant improvements. Actually, as can be seen from the detailed results, both SST$_{train}$ and ERC$_{true}$ perform worse than ST in several cases. This issue is studied in more detail in the following subsection.

Perhaps even more interestingly, \textit{none of the state-of-the-art multi-target methods participating in this comparison manages to significantly improve the performance of ST}. In fact, ST is ranked second after SST$_{train}$ in the per dataset analysis and third after SST$_{train}$ and RLC in the per target analysis, and is found significantly better than TNR and \textsc{Dirty} in both types of analyses.
This exceptionally good performance of ST might seem a bit surprising given the results of previous studies \citep[e.g.][]{kocev2007,tsoumakas:2014:ecmlpkdd} but is in accordance with empirical and theoretical results for Binary Relevance (as discussed in Section~\ref{sec:methods}) and is attributed to the use of a very strong base regressor.

\begin{figure}
\centering
    \begin{subfigure}[b]{0.49\textwidth}
    	\resizebox{1\textwidth}{!}{
        \includegraphics[trim = 4.5cm 10.8cm 3.2cm 10cm, clip]{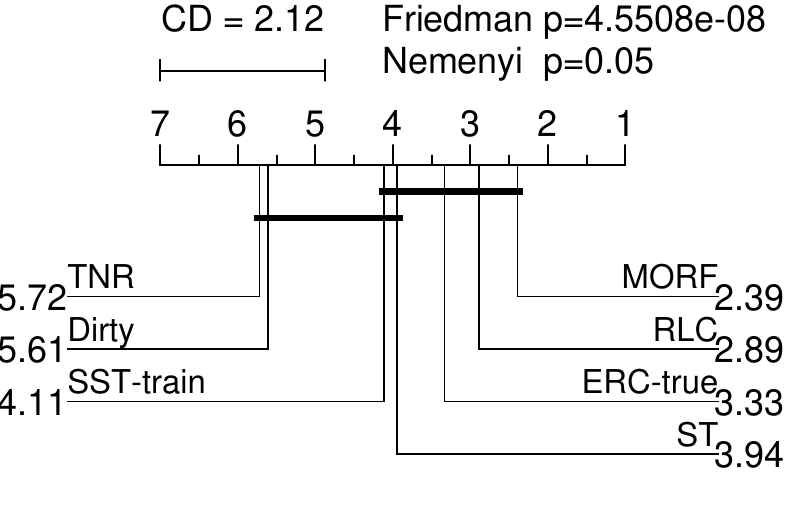}
        }
        \caption{Per dataset analysis.}
    \end{subfigure}
    ~
    \begin{subfigure}[b]{0.49\textwidth}
        \resizebox{1\textwidth}{!}{
        \includegraphics[trim = 4.5cm 10.8cm 3.2cm 10cm, clip]{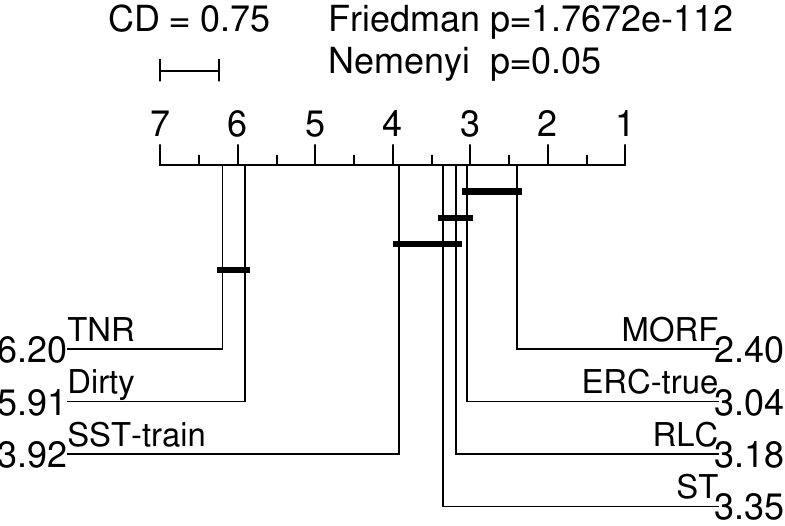}
        }
        \caption{Per target analysis.}       
    \end{subfigure}
    \caption{Comparison of direct adaptations (with \textsc{ridge} as base regressor) using the Nemenyi test. Groups of methods that are not significantly different (at $p=0.05$) are connected.}
	\label{fig:nemenyi:straightforward:ridge}
\end{figure}

To validate this, we instantiate all problem transformation methods with \textsc{ridge}, a base regressor that was found to perform worse than \textsc{bag} in Subsection~\ref{sec:results:base}, and repeat the comparison.
As shown in Figure~\ref{fig:nemenyi:straightforward:ridge}, the situation is quite different compared to when \textsc{bag} was used as base regressor.
We observe that ST is now ranked below MORF, RLC and ERC$_{true}$ in both the per dataset and the per target analysis and is found significantly worse than MORF according to the per target analysis.
Clearly, \textit{as the strength of the base regressor increases, i.e. when the information provided by the features is well exploited, improving the performance of ST becomes more difficult}. However, it is this challenging setting where performance improvements matter the most and it is thus interesting to see whether the proposed extensions of SST and ERC manage to obtain more consistent improvements over ST (compared to SST$_{train}$ and ERC$_{true}$) under this setting.

\subsection{Evaluation of Meta-input Generation Variants}
\label{sec:results:variants}

In this subsection we evaluate the performance of SST and ERC when different types of values are used for the meta-inputs at training time. In particular, each method is evaluated using the actual target values ($true$ variants), in-sample estimates ($train$ variants) and out-of-sample estimates ($cv$ variants) generated using the proposed internal cross-validation strategy. We want to see whether the $cv$ variants (that according to the discussion of Subsection~\ref{sec:methods:meta} are expected to be less affected by the discrepancy problem) can indeed perform better than the $train$ and $true$ variants and whether they manage to obtain more consistent improvements over ST. We also want to see how the SST variants compare to the ERC variants.

\begin{figure}
\centering
    \begin{subfigure}[b]{0.49\textwidth}
    	\resizebox{1\textwidth}{!}{
        \includegraphics[trim = 4.5cm 13.9cm 3.2cm 10cm, clip]{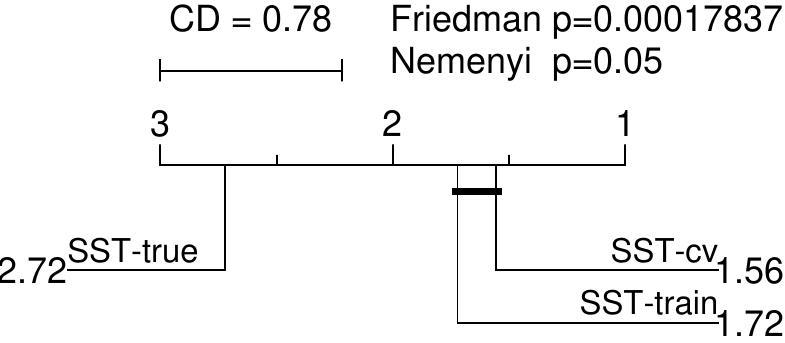}
        }
        \caption{SST, per dataset analysis.}
    \end{subfigure}
    ~
    \begin{subfigure}[b]{0.49\textwidth}
        \resizebox{1\textwidth}{!}{
        \includegraphics[trim = 4.5cm 13.9cm 3.2cm 10cm, clip]{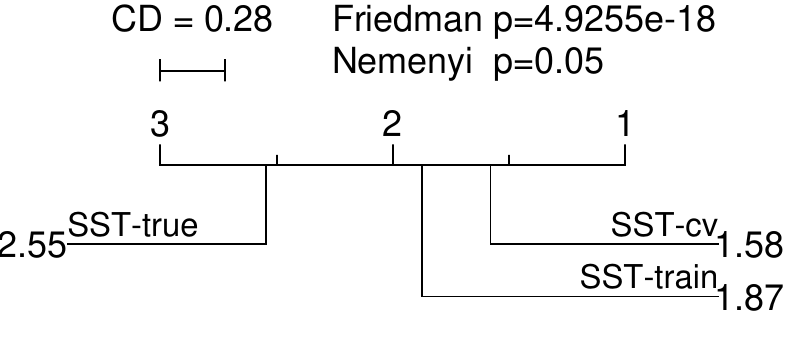}
        }
        \caption{SST, per target analysis.}      
    \end{subfigure}
    \caption{Comparison of SST variants using the Nemenyi test. Groups of methods that are not significantly different (at $p=0.05$) are connected.}
	\label{fig:nemenyi:variants_sst}
\end{figure}

\begin{figure}
\centering
    \begin{subfigure}[b]{0.49\textwidth}
    	\resizebox{1\textwidth}{!}{
        \includegraphics[trim = 4.5cm 13.9cm 3.2cm 10cm, clip]{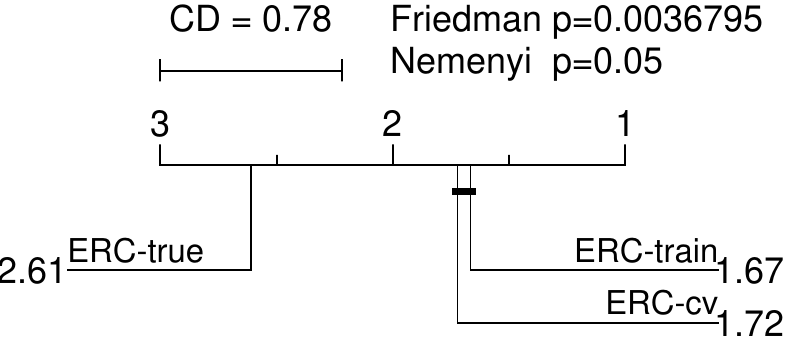}
        }
        \caption{ERC, per dataset analysis.}
    \end{subfigure}
    ~
    \begin{subfigure}[b]{0.49\textwidth}
        \resizebox{1\textwidth}{!}{
        \includegraphics[trim = 4.5cm 13.9cm 3.2cm 10cm, clip]{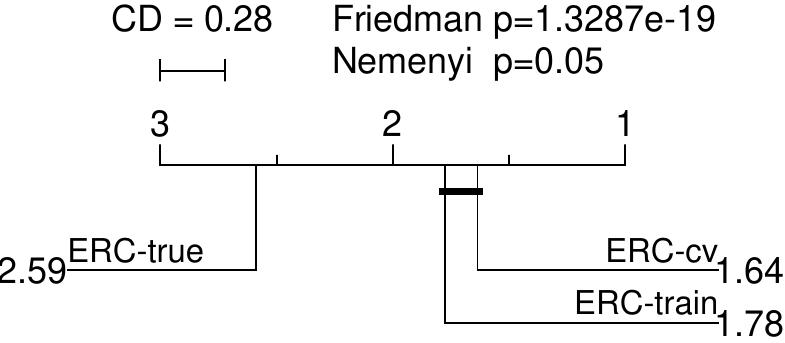}
        }
        \caption{ERC, per target analysis.}      
    \end{subfigure}
    \caption{Comparison of ERC variants using the Nemenyi test. Groups of methods that are not significantly different (at $p=0.05$) are connected.}
	\label{fig:nemenyi:variants_erc}
\end{figure}

Figures~\ref{fig:nemenyi:variants_sst} and \ref{fig:nemenyi:variants_erc} show the average ranks and the results of the Friedman and Nemenyi tests for the three variants of SST and ERC, respectively, according to the per dataset (left) and the per target (right) analysis.
First, we see that in both SST and ERC and in both types of analyses, the variants that use the actual values of the targets ($true$) obtain the worst average ranks and are found significantly worse than both variants that use estimates ($train$ and $cv$). 
Since the variants of each method differ only with respect to the type of values that they use for the meta-inputs, it is clear that \textit{the discrepancy problem has a significant impact on the performance of both SST and ERC and that the use of estimates can ameliorate this problem}.

With respect to the kind of estimates that should be used (in-sample or out-of-sample) the situation is slightly different for each method. 
In the case of SST, the $cv$ variant obtains the best average rank in both the per dataset and the per target analysis and its difference with the $train$ variant is found significant in the per target analysis.
In the case of ERC, while the $cv$ variant is ranked higher than the $train$ variant in the per target analysis, the $train$ variant is ranked slightly higher in the per dataset analysis and in both cases the differences are not found significant. 
This suggests that \textit{using out-of-sample estimates is important for SST while ERC seems to be less affected by the discrepancy problem and, as a result, the use of in-sample estimates can be considered as a viable alternative}.

\begin{figure}
\centering
    \begin{subfigure}[b]{0.49\textwidth}
    	\resizebox{1\textwidth}{!}{
        \includegraphics[trim = 4.5cm 11.0cm 3.2cm 10cm, clip]{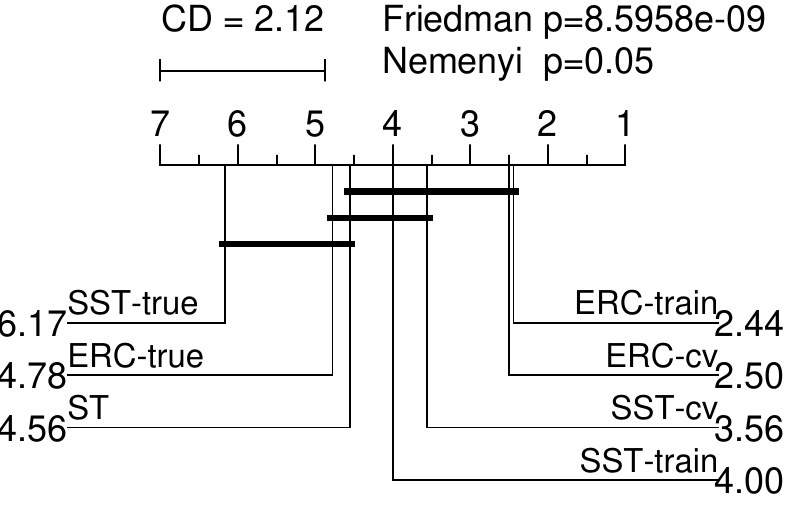}
        }
        \caption{Per dataset analysis.}
    \end{subfigure}
    ~
    \begin{subfigure}[b]{0.49\textwidth}
        \resizebox{1\textwidth}{!}{
        \includegraphics[trim = 4.5cm 11.0cm 3.2cm 10cm, clip]{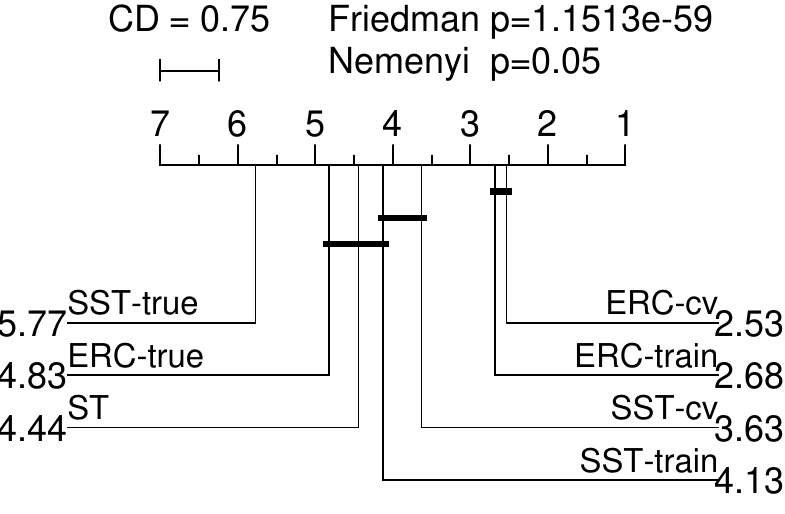}
        }
        \caption{Per target analysis.}        
    \end{subfigure}
    \caption{Comparison of SST$_{true/train/cv}$, ERC$_{true/train/cv}$ and ST using the Nemenyi test. Groups of methods that are not significantly different (at $p=0.05$) are connected.}
	\label{fig:nemenyi:variants_all}
\end{figure}

A question that has not been answered yet, is how the new variants of SST and ERC compare to ST and to each other. Figure~\ref{fig:nemenyi:variants_all} shows the results of the Friedman and the Nemenyi tests when all variants of SST and ERC are compared together with ST.
We see that in both the per dataset (left) and the per target (right) analysis, the four variants that use estimates for the meta-inputs obtain lower average ranks than ST while the $true$ variants obtain worse average ranks. The differences with ST are not found significant according to the per dataset analysis but according to the per target analysis ERC$_{train}$ and ERC$_{cv}$ are found significantly better.
Comparing the SST variants with the ERC variants, we see that each ERC variant is always ranked above the corresponding SST variant. This suggests that \textit{ERC's strategy for leveraging information from target variables is beneficial}.
Moreover, we see that that ERC$_{train}$ and ERC$_{cv}$ \textit{are found significantly better than the rest of the methods} according to the per target analysis.

\subsubsection{Cautiousness Analysis}

So far, our analysis has focused on the average performance of the proposed methods (as quantified by their average ranks over datasets and targets) and we found that ERC$_{train}$ and ERC$_{cv}$ outperform the independent regressions baseline significantly.
However, it is also important to see the consistency of these improvements across different datasets and targets. 
In particular, we would like to study the degree of cautiousness that each method exhibits, i.e. how frequently and to what extent are the predictions produced by each method less accurate than the predictions of ST.

\begin{figure}
\centering
    \begin{subfigure}[b]{0.90\textwidth}
    	\resizebox{1\textwidth}{!}{
        \includegraphics[trim = 1.8cm 7.1cm 1.9cm 9.1cm, clip]{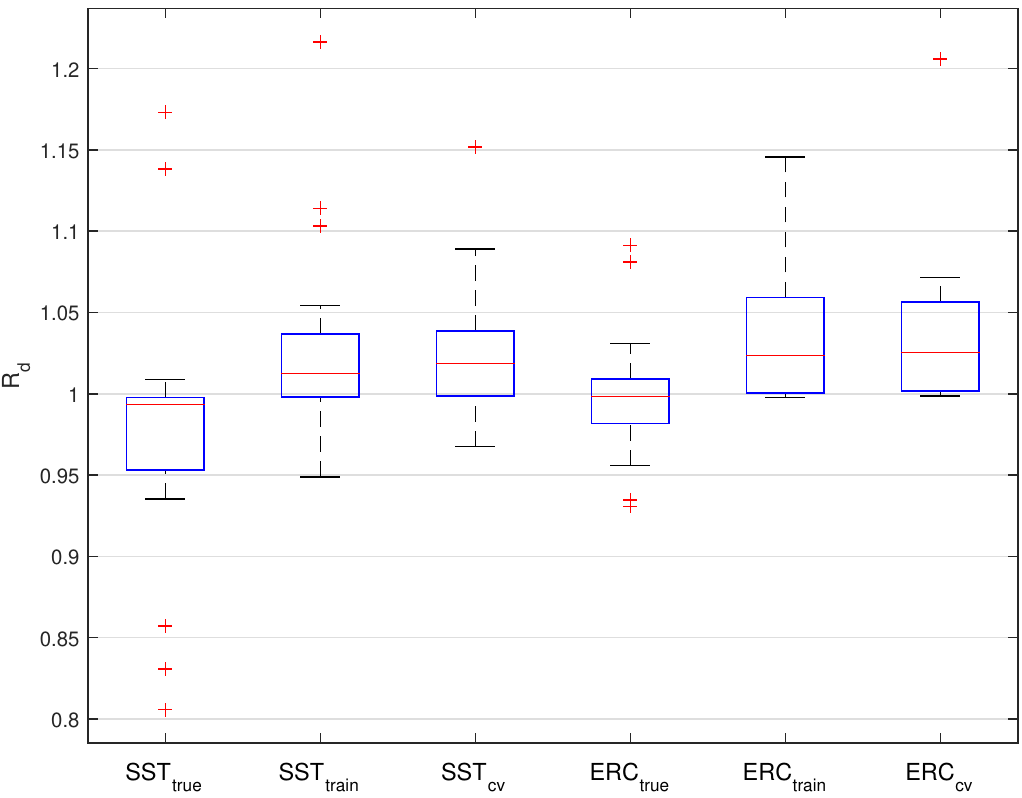}
        }
        \caption{Distribution of $R_d$ values for each method over 18 datasets.}
    \end{subfigure}
    \begin{subfigure}[b]{0.90\textwidth}
        \resizebox{1\textwidth}{!}{
        \includegraphics[trim = 1.8cm 7.1cm 1.9cm 9.1cm, clip]{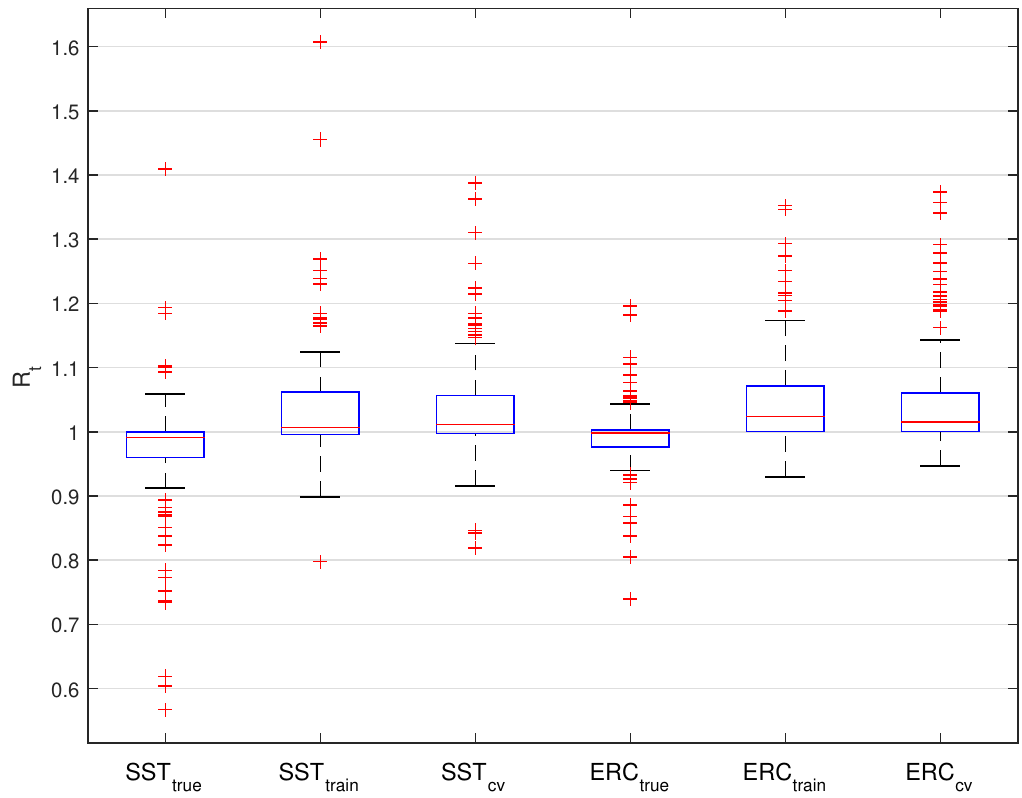}
        }
        \caption{Distribution of $R_t$ values for each method over 143 targets.}
    \end{subfigure}
    \caption{Cautiousness analysis of SST and ERC variants. On each box, the whiskers extend to the most extreme data points still within 1.5 IQR of the lower quartile and outliers are plotted individually.}
	\label{fig:boxplots:variants}
\end{figure}

To facilitate a comparison of the methods in this regard, the following measures are defined:\\
\noindent\begin{minipage}{.5\linewidth}
\begin{equation*}
  R_d(M) = \frac{aRRMSE(ST)}{aRRMSE(M)},
\end{equation*}
\end{minipage}%
\begin{minipage}{.5\linewidth}
\begin{equation*}
  R_t(M) = \frac{RRMSE(ST)}{RRMSE(M)}.
\end{equation*}
\end{minipage}
For each method $M$ and dataset $d$, $R_d$ quantifies the amount of improvement or degradation induced by $M$ compared to ST in terms of $aRRMSE$.  
Similarly, for each method $M$ and target $t$, $R_t$ quantifies the amount of improvement or degradation compared to ST in terms of $RRMSE$.
Values of $R_d(M)$ and $R_t(M)$ $<1$ indicate that the method produces worse estimates than ST ($R_d(ST)=R_t(ST)=1$).
The upper part of Figure~\ref{fig:boxplots:variants} displays box plots of the values of $R_d$ over the 18 datasets included in the experimental study, i.e. each box plot summarizes the distribution of 18 values, while the lower part displays box plots of the values of $R_t$ over 143 targets, i.e. each box plot summarizes the distribution of 143 values.

We see that that in both the per dataset and the per target analysis, the $true$ variants are the ones exhibiting the more dispersed distributions with several cases of significant degradation of ST's performance.
The $train$ and $cv$ variants are clearly more cautious with much fewer cases of degradation and even fewer cases of significant degradation. 
Looking at the distributions of $R_t$, we could say that \textit{the $cv$ variants appear a bit more cautious than the $train$ variants especially in the case of SST}.
We also see that the ERC variants are always more cautious than the corresponding SST variants.
Clearly, ERC$_{train}$ and ERC$_{cv}$ are the two most cautious methods since they obtain very similar or better performance than ST on all datasets and on about 75\% of the targets. Even on targets where the two methods obtain a lower performance than ST, the reduction is less than about 5\%. 
This characteristic along with the fact that they obtain the largest average improvements over ST, make ERC$_{train}$ and ERC$_{cv}$ highly appealing.

\subsection{Comparison with the State-of-the-art}
\label{sec:results:soa}

\begin{figure}
\centering
    \begin{subfigure}[b]{0.49\textwidth}
    	\resizebox{1\textwidth}{!}{
        \includegraphics[trim = 4.5cm 10.8cm 3.2cm 10cm, clip]{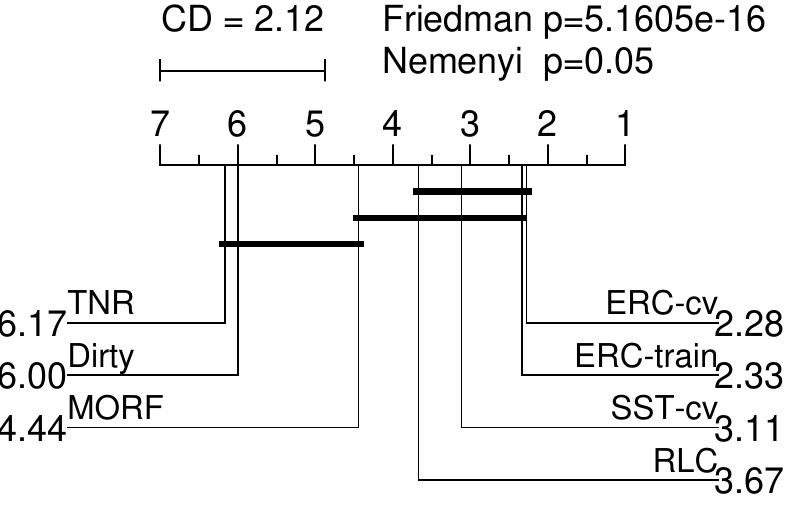}
        }
        \caption{Per dataset analysis.}
    \end{subfigure}
    ~
    \begin{subfigure}[b]{0.49\textwidth}
        \resizebox{1\textwidth}{!}{
        \includegraphics[trim = 4.5cm 10.8cm 3.2cm 10cm, clip]{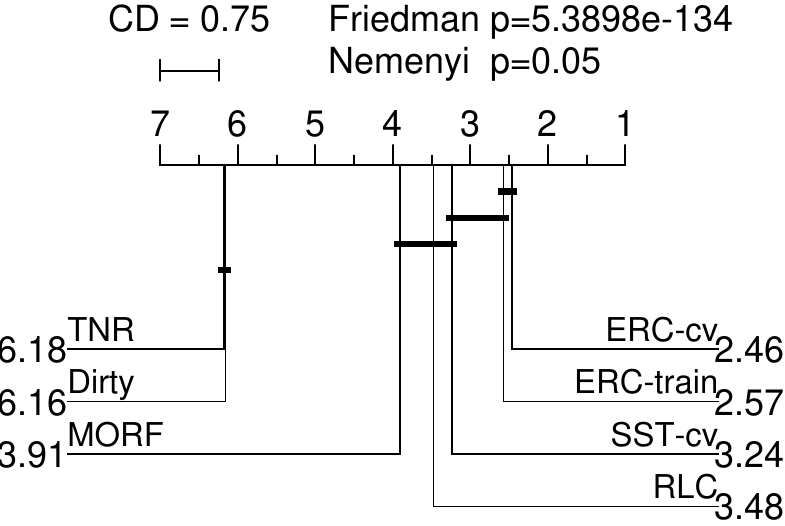}
        }
        \caption{Per target analysis.}      
    \end{subfigure}
    \caption{Comparison of the best SST and ERC variants with the state-of-the-art using the Nemenyi test. Groups of methods that are not significantly different (at $p=0.05$) are connected.}
	\label{fig:nemenyi:soa}
\end{figure}

\begin{figure}
\centering
    \begin{subfigure}[b]{0.9\textwidth}
    	\resizebox{1\textwidth}{!}{
        \includegraphics[trim = 1.8cm 7.1cm 1.9cm 9.1cm, clip]{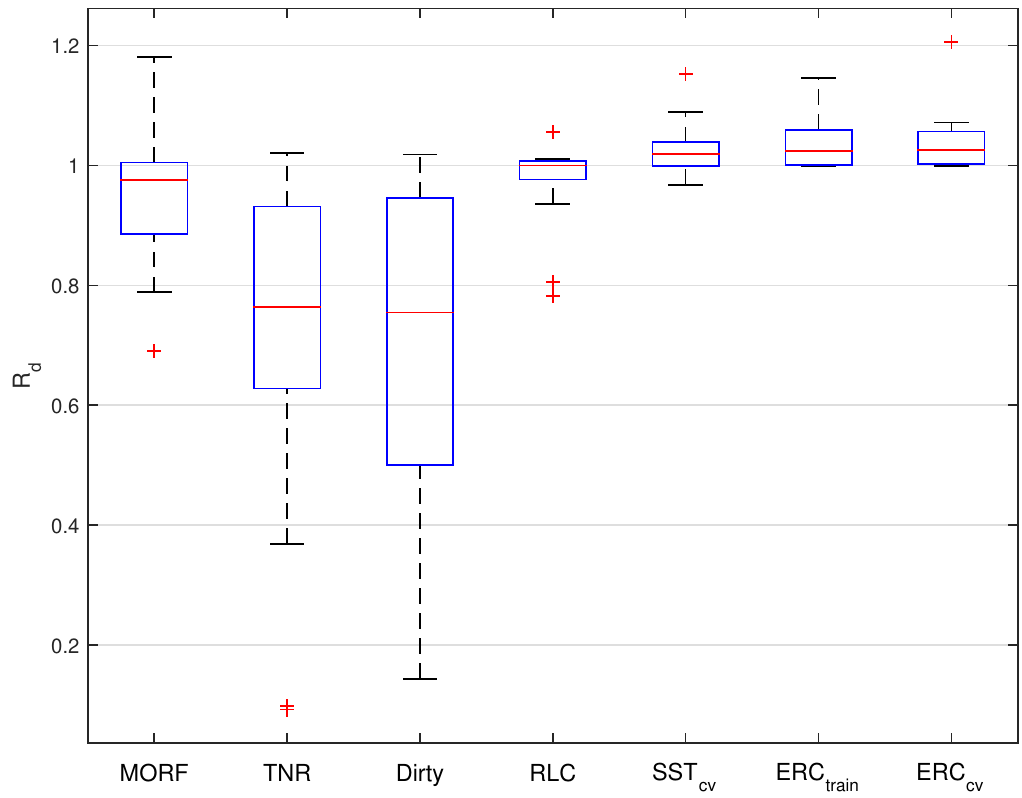}
        }
        \caption{Distribution of $R_d$ values for each method over 18 datasets.}
    \end{subfigure}
    \begin{subfigure}[b]{0.9\textwidth}
        \resizebox{1\textwidth}{!}{
        \includegraphics[trim = 1.8cm 7.1cm 1.9cm 9.1cm, clip]{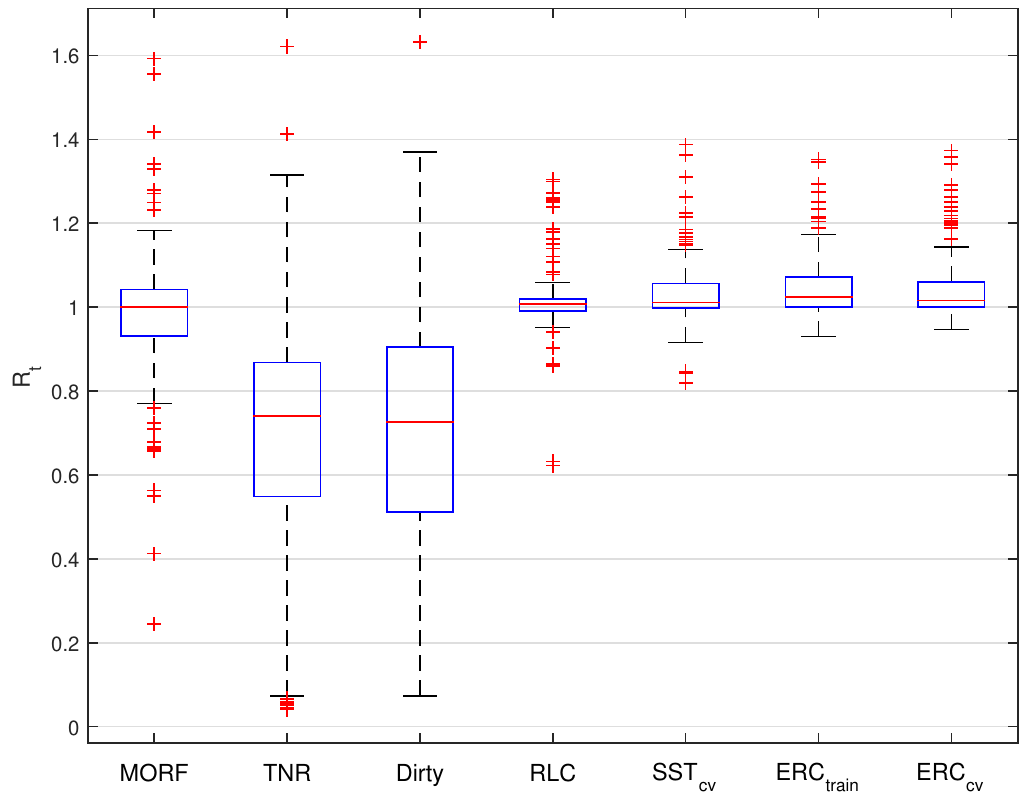}
        }
        \caption{Distribution of $R_t$ values for each method over 143 targets.}
    \end{subfigure}
    \caption{Comparing the cautiousness of the best SST and ERC variants to that of state-of-the-art methods. On each box, the whiskers extend to the most extreme data points still within 1.5 IQR of the lower quartile and outliers are plotted individually.}
	\label{fig:boxplots:soa}
\end{figure}

In this section we compare the three best performing variants of the proposed methods, i.e. ERC$_{cv}$, ERC$_{train}$ and SST$_{cv}$, with MORF, RLC, TNR and \textsc{Dirty} to see how they compare to the state-of-the-art. Figure~\ref{fig:nemenyi:soa} shows the results of the Friedman and Nemenyi tests for the analysis per dataset (left) and per target (right). The per dataset analysis shows that all our methods perform significantly better than TNR and \textsc{Dirty} while ERC$_{cv}$ and ERC$_{train}$ also perform significantly better than MORF. Moreover, all our methods obtain a lower average rank than RLC but according to this analysis the differences are not significant.
According to the per target analysis, all our methods are found significantly better than TNR, \textsc{Dirty} and MORF, and additionally, ERC$_{cv}$ and ERC$_{train}$ are found significantly better than RLC.
In Figure~\ref{fig:boxplots:soa} we compare the performance of the methods from a cautiousness perspective, as we did in Subsection~\ref{sec:results:variants}. TNR, \textsc{Dirty} and MORF are far less cautious than SST$_{cv}$, ERC$_{train}$ and ERC$_{cv}$ with many instances of extreme degradation of ST's performance.
RLC is more cautious but not as much as SST$_{cv}$, ERC$_{train}$ and ERC$_{cv}$, especially according to the per target analysis.

\subsection{Running Times}
\label{sec:results:runtimes}

In this subsection we compare the running times of the studied methods. 
Experiments were run on a 64-bit CentOS Linux machine with 80 Intel Xeon E7-4860 processors running at 2.27 GHz and 1 TB of main memory. 
The detailed results per method and dataset are shown in Table~\ref{tbl:times}.
For ST, RLC, SST and ERC we report times with \textsc{bag}
as base regressor. 
The number shown in parenthesis next to the name of each dataset corresponds to the maximum number of processor threads that were available during the experiment. ST, SST, ERC and RLC made use of multiple threads through Weka's multi-threaded implementation of Bagging. 
Thus, running times are directly comparable for these methods.
Multi-threading was also partly used in TNR for the computation of the gradients. \textsc{dirty} and MORF, on the other hand, always used a single processor thread.

Looking at the aggregated running times, we see that MORF is by far the most efficient method, followed by ST, SST$_{true}$ and SST$_{train}$ which have similar running times. On the other hand, \textsc{dirty} is the least efficient method, followed by ERC$_{cv}$. The running times of the rest of the methods lie in between.
With respect to the SST and ERC variants, we see that their running times agree with the complexity analysis of Subsection~\ref{sec:methods:complexity}. The total running time of SST$_{true}$ is roughly twice the total running time of ST and similar to the total running time of SST$_{train}$. SST$_{cv}$ is the least efficient among SST variants with a total running time that is about 5 times larger than that of SST$_{true}$ and SST$_{train}$. With respect to the ERC variants, we see that ERC$_{true}$ and ERC$_{train}$ have similar total running times (which are also roughly similar to the total running time of SST$_{cv}$) while ERC$_{cv}$ is about 7.5 times slower. 

Overall, we see that the improvements achieved by ERC$_{cv}$ and ERC$_{train}$ over ST come with an increased computational cost. However, this cost is manageable especially in the case of ERC$_{train}$. Furthermore, when better efficiency is needed, besides the use of parallelization one might consider reducing the ensemble size ($k$) or using a smaller number of folds ($f$) when applying internal cross-validation (in ERC$_{cv}$).

\renewcommand{\tabcolsep}{2pt} 

\begin{table}
\caption{Running times (in seconds) using \textsc{bag} as base regressor. The number in parenthesis next to the name of each dataset corresponds to the maximum number of processor threads that could be utilized during the experiment.}
\label{tbl:times}
\begin{center}
\scriptsize
\begin{tabular}{l | *{11}{r}}
\hline
Dataset  & \multicolumn{1}{r}{\rotatebox{90}{ST}} & \multicolumn{1}{r}{\rotatebox{90}{SST$_{true}$}} & \multicolumn{1}{r}{\rotatebox{90}{SST$_{train}$}} & \multicolumn{1}{r}{\rotatebox{90}{SST$_{cv}$}} & \multicolumn{1}{r}{\rotatebox{90}{ERC$_{true}$}} & \multicolumn{1}{r}{\rotatebox{90}{ERC$_{train}$}} & \multicolumn{1}{r}{\rotatebox{90}{ERC$_{cv}$}} & \multicolumn{1}{r}{\rotatebox{90}{MORF}} & \multicolumn{1}{r}{\rotatebox{90}{RLC}} & \multicolumn{1}{r}{\rotatebox{90}{TNR}} & \multicolumn{1}{r}{\rotatebox{90}{\textsc{dirty}}}\\\hline
edm (2) & 4.3 & 3.3 & 3.2 & 14.5 & \textbf{2.8} & 2.8 & 13.7 & 6.3 & 78.3 & 16.6 & 281.4 \\
sf1 (3) & 4.1 & \textbf{4.1} & 4.3 & 16.7 & 11.2 & 11.8 & 60.8 & 8.1 & 79.2 & 11.8 & 60.5 \\
sf2 (3) & \textbf{7.5} & 11.5 & 11.4 & 47.6 & 33.2 & 36.3 & 262.0 & 15.4 & 238.7 & 17.3 & 56.9 \\
jura (3) & \textbf{10.9} & 15.3 & 14.9 & 72.6 & 46.4 & 46.9 & 264.2 & 14.0 & 199.1 & 17.0 & 49.2 \\
wq (6) & \textbf{43.3} & 104.0 & 141.7 & 540.0 & 561.3 & 607.0 & 4648.4 & 106.3 & 295.0 & 143.7 & 531.2 \\
enb (2) & \textbf{10.4} & 15.4 & 15.9 & 76.0 & 16.7 & 16.4 & 74.9 & 15.7 & 313.9 & 35.8 & 223.0 \\
slump (3) & 4.7 & 4.2 & \textbf{4.0} & 16.5 & 11.4 & 11.1 & 67.8 & 4.2 & 48.5 & 10.7 & 55.6 \\
andro (2) & 8.2 & 10.6 & 9.0 & 47.9 & 49.5 & 47.2 & 360.3 & \textbf{3.1} & 72.5 & 445.3 & 1670.8 \\
onsales (8) & 317.6 & 612.2 & 560.8 & 2803.0 & 2835.0 & 2801.6 & 21568.6 & \textbf{43.2} & 1628.0 & 2769.1 & 80616.2 \\
scpf (3) & 22.9 & 36.2 & 36.6 & 196.5 & 106.2 & 106.4 & 651.2 & \textbf{13.4} & 491.3 & 84.5 & 215.4 \\
atp1d (4) & 155.6 & 356.1 & 345.7 & 1614.7 & 1496.0 & 1555.6 & 12169.2 & \textbf{23.3} & 2408.0 & 2212.8 & 57276.5 \\
atp7d (4) & 145.8 & 318.6 & 313.9 & 1517.2 & 1336.9 & 1385.0 & 11158.2 & \textbf{18.4} & 2179.2 & 1536.0 & 53974.0 \\
oes97 (6) & 199.5 & 454.8 & 422.2 & 1998.4 & 1977.3 & 1965.5 & 16609.1 & \textbf{32.7} & 1057.1 & 4581.3 & 124231.5 \\
oes10 (6) & 286.4 & 613.1 & 535.7 & 2725.9 & 2691.0 & 2575.1 & 21446.6 & \textbf{39.4} & 1182.6 & 6093.8 & 157399.6 \\
rf1 (8) & 379.6 & 868.2 & 890.8 & 4180.7 & 3885.7 & 4018.8 & 30050.1 & \textbf{351.1} & 4184.7 & 2952.1 & 34927.2 \\
rf2 (10)& 3505.5 & 6539.0 & 6065.6 & 31562.8 & 28347.1 & 28104.0 & 234088.5 & \textbf{325.2} & 35429.1 & 18130.9 & 197482.1 \\
scm1d (10) & 1398.5 & 2499.5 & 2398.3 & 11111.9 & 10465.6 & 11253.2 & 113500.0 & \textbf{199.7} & 6018.6 & 5647.1 & 105215.8 \\
scm20d (8) & 302.6 & 613.9 & 621.8 & 2615.8 & 2733.1 & 2646.9 & 18549.2 & \textbf{140.8} & 1627.4 & 1120.2 & 4779.6 \\
\hline
\hline
Total  & 6807.6 & 13079.8 & 12395.8 & 61158.7 & 56606.3 & 57191.6 & 485542.9 & \textbf{1360.2} & 57531.3 & 45826.0 & 819046.5 \\
\hline
\end{tabular}
\end{center}
\end{table}

\renewcommand{\tabcolsep}{6pt} 

\subsection{Discussion}
\label{sec:results:discussion}

Several interesting conclusions can be drawn from our experimental results. The experiments of Subsections~\ref{sec:results:straightforward} and \ref{sec:results:variants} showed that while the directly adapted versions of SST and ERC have comparable or better performance than state-of-the-art methods, a careful handling of the discrepancy problem is crucial for obtaining consistent improvements over the independent regressions baseline and the state-of-the-art.
In particular, as the experiments of Subsection~\ref{sec:results:variants} revealed, the use of estimates for the meta-inputs during training should clearly be preferred over using the actual target values. 
With regard to using in-sample versus out-of-sample estimates, the results indicate that while out-of-sample estimates are preferable in SST, ERC performs almost equally well using either type of estimates for the meta-inputs. As discussed in Subsection~\ref{sec:methods:discussion}, ERC's models are built on input spaces which are expanded with fewer meta-inputs compared to SST's models and, as a result, a smaller amount of error accumulation is risked at prediction time.

Another interesting conclusion is that when a strong base regressor is employed, the task of improving the performance of ST becomes very difficult. As a result, multi-target methods which are considered state-of-the-art fail to improve ST's performance and are even performing significantly worse. This was particularly the case for the two multi-task methods, TNR and \textsc{Dirty}, which were consistently found to be the worst performers.
One explanation for their bad performance is the fact that both methods are based on a linear formulation of the problem that, as revealed by the base regressor exploration experiments, is not the most suitable hypothesis representation for the studied datasets (\textsc{ridge} and \textsc{svr} performed worse than \textsc{sgb} and \textsc{bag} that are based on a non-linear hypothesis representation).
Moreover, multi-task methods are expected to work better than single-task methods in cases where there is a lack of training data for some of the tasks \citep{alvarez2011kernels}. This is not the case for most of the datasets that we used in this study as well as many recent multi-target prediction problems. In fact, the two datasets where TNR and \textsc{Dirty} perform better than ST (sf1 and slump) are among those with the fewest training examples.

With respect to MORF, although it was found significantly more competitive than TNR and \textsc{Dirty}, it also performed worse than ST on average. Nevertheless, we should point out that MORF achieved the best accuracy on three datasets (edm, wq, andro) and is the most computationally efficient of the compared methods. Similarly to TNR and \textsc{Dirty}, MORF has the disadvantage of having a fixed hypothesis representation (trees), as opposed to the proposed methods that have the ability of adapting better to a specific domain by being instantiated with a more suitable base regressor. This advantage of the proposed methods is shared with RLC which, however, was not found as accurate. 

Overall, our experimental results demonstrate that of the methods proposed in this paper, ERC$_{train}$ and ERC$_{cv}$ and, to a lesser extent, SST$_{train}$ and SST$_{cv}$ provide increased accuracy over doing a separate regression per target. In addition, ERC$_{train}$ and ERC$_{cv}$ are significantly more accurate than TNR, \textsc{Dirty}, MORF and RLC (in the per target analysis).
If caution is a further concern, then again ERC$_{train}$ and ERC$_{cv}$ compare favorably to the rest of the methods.
With respect to the $true$ variants of SST and ERC, we should stress out that despite having a worse average performance, they are worthy of being considered by a practitioner as they obtain the highest performance in datasets (e.g., sf1 and scfp) where the discrepancy problem is not predominant.
\section{Conclusion}
\label{sec:conclusions}

Motivated by the similarity between the tasks of multi-label classification and multi-target regression, this paper introduced SST and ERC, two new multi-target regression techniques derived through a simple adaptation of two well-known multi-label classification methods. Both methods are based on the idea of treating other prediction targets as additional input variables, and represent a conceptually simple way of exploiting target dependencies in order to improve prediction accuracy.

A comprehensive experimental analysis that includes a multitude of real-world datasets and four existing state-of-the-art methods, reveals that, despite being competitive with the state-of-the-art, the directly adapted versions of SST and ERC do not manage to obtain significant improvements or even degrade the performance of the independent regressions baseline. 
This degradation is attributed to an underestimation (in the original formulations of the methods) of the impact of the discrepancy of the values used for the additional input variables between training and prediction.
Confirming our hypothesis, extended versions of the methods that attempt to mitigate the discrepancy using out-of-sample estimates of the targets during training, manage to obtain consistent and significant improvements over the baseline approach and are found significantly better than four state-of-the-art methods.
The fact that these impressive results were obtained by applying relatively simple adaptations of existing multi-label classification methods, highlights the importance of exploiting relationships between similar machine learning tasks.

Concluding, let us point to some directions for future work.
Although a mitigation of the discrepancy problem leads to significant performance improvements, a different amount of mitigation is ideal for each target. As a result, the use of in-sample estimates (or even the actual target values) gives better results for some targets. 
Thus, a promising direction for future work would be a deeper theoretical analysis of the different variants and the identification of problem characteristics that favor the use of one variant over the other.
Finally, we should point out that SST and ERC can be viewed as strategies for leveraging variables that are available in the training phase but not in the prediction phase. This type of scenario is very common, for instance, in time series prediction. We believe that adapting SST and ERC for this type of problems is another valuable opportunity for future work.


\bibliographystyle{spbasic}      
\bibliography{citations}   

\appendix
\section*{Appendix}
\label{appendix}

\section{Datasets}
\label{appendix:datasets}

\subsection{Existing Datasets}
\label{appendix:datasets:existing}

\noindent \textbf{EDM}
The Electrical Discharge Machining dataset \citep{karalic1997} represents a two-target regression problem. The task is to shorten the machining time by reproducing the behaviour of a human operator that controls the values of two variables. Each of the target variables takes 3 distinct numeric values ($\{-1,0,1\}$) and there are 16 continuous input variables.

\noindent \textbf{SF}
The Solar Flare dataset \citep{Lichman:2013} has 3 target variables that correspond to the number of times 3 types of solar flare (common, moderate, severe) are observed within 24 hours. There are two versions of this dataset. SF1 contains data from year 1969 and SF2 from year 1978.

\noindent \textbf{JURA}
The Jura
\citep{goovaerts1997} dataset consists of measurements of concentrations of seven heavy metals (cadmium, cobalt, chromium, copper, nickel, lead, and zinc), recorded at 359 locations in the topsoil of a region of the Swiss Jura. The type of land use (Forest, Pasture, Meadow, Tillage) and rock type (Argovian, Kimmeridgian, Sequanian, Portlandian, Quaternary) were also recorded for each location. In a typical scenario \citep{goovaerts1997,alvarez2011}, we are interested in the prediction of the concentration of metals that are more expensive to measure (primary variables) using measurements of metals that are cheaper to sample (secondary variables). In this study, cadmium, copper and lead are treated as target variables while the remaining metals along with land use type, rock type and the coordinates of each location are used as predictive features.

\noindent \textbf{WQ}
The Water Quality dataset \citep{dzeroski2000} has 14 target attributes that refer to the relative representation of plant and animal species in Slovenian rivers and 16 input attributes that refer to physical and chemical water quality parameters.

\subsection{New Datasets}
\label{appendix:datasets:new}
	
\noindent \textbf{ENB}
The Energy Building dataset \citep{tsanas2012accurate} concerns the prediction of the heating load and cooling load requirements of buildings (i.e. energy efficiency) as a function of eight building parameters such as glazing area, roof area, and overall height, amongst others. 

\noindent \noindent \textbf{SLUMP}
The Concrete Slump dataset \citep{Yeh2007474} concerns the prediction of three properties of concrete (slump, flow and compressive strength) as a function of the content of seven concrete ingredients: cement, fly ash, blast furnace slag, water, superplasticizer, coarse aggregate, and fine aggregate. 

\noindent \textbf{ANDRO}
The Andromeda dataset \citep{hatzikos2008} concerns the prediction of future values for six water quality variables (temperature, pH, conductivity, salinity, oxygen, turbidity) in Thermaikos Gulf of Thessaloniki, Greece. Measurements of the target variables are taken from under-water sensors with a sampling interval of 9 seconds and then averaged to get a single measurement for each variable over each day. The specific dataset that we use here corresponds to using a window of 5 days (i.e. features attributes correspond to the values of the six water quality variables up to 5 days in the past) and a lead of 5 days (i.e. we predict the values of each variable 6 days ahead). 

\noindent \textbf{OSALES}
This is a pre-processed version of the dataset used in Kaggle's ``Online Product Sales'' competition \citep{kaggle:onsales} that concerns the prediction of the online sales of consumer products. Each row in the dataset corresponds to a different product that is described by various product features as well as features of an advertising campaign. There are 12 target variables corresponding to the monthly sales for the first 12 months after the product launches. For the purposes of this study we removed examples with missing values in any target variable (112 out of 751) and attributes with one distinct value (145 out of 558).

\noindent \textbf{SCPF}
This is a pre-processed version of the dataset used in Kaggle's ``See Click Predict Fix'' competition \citep{kaggle:scpf}. It concerns the prediction of three target variables that represent the number of views, clicks and comments that a specific 311 issue will receive. The issues have been collected from 4 cities (Oakland, Richmond, New Haven, Chicago) in the US and span a period of 12 months (01/2012 - 12/2012). The version of the dataset that we use here is a random 1\% sample of the data. 
In terms of features we use the number of days that an issues stayed online, the source from where the issue was created (e.g. android, iphone, remote api, etc.), the type of the issue (e.g. graffiti, pothole, trash, etc.), the geographical co-ordinates of the issue, the city it was published from and the distance from the city center. All multi-valued nominal variables were first transformed to binary and then rare binary variables (being true for less than 1\% of the cases) were removed. 

\noindent \textbf{OES}
The Occupational Employment Survey datasets were obtained from years 1997 (OES97) and 2010 (OES10) of the annual Occupational Employment Survey compiled by the US Bureau of Labor Statistics.
Each row provides the estimated number of full-time equivalent employees across many employment types for a specific metropolitan area. There are 334 and 403 cities in the 1997 and 2010 datasets, respectively. The input variables in these datasets are a randomly sequenced subset of employment types (e.g. doctor, dentist, car repair technician, etc.) observed in at least 50\% of the cities (some categories had no values for particular cities). The targets for both years are randomly selected from the entire set of categories above the 50\% threshold. Missing values in both the input and the target variables were replaced by sample means for these results. To our knowledge, this is the first use of the OES dataset for benchmarking of multi-target prediction algorithms.

\noindent \textbf{ATP}
The Airline Ticket Price dataset concerns the prediction of airline ticket prices. The rows are a sequence of time-ordered observations over several days. Each sample in this dataset represents a set of observations from a specific observation date and departure date pair. The input variables for each sample are values that may be useful for prediction of the airline ticket prices for a specific departure date. The target variables in these datasets are the next day (ATP1D) price or minimum price observed over the next 7 days (ATP7D) for 6 target flight preferences: 1) any airline with any number of stops, 2) any airline non-stop only, 3) Delta Airlines, 4) Continental Airlines, 5) Airtrain Airlines, and 6) United Airlines. The input variables include the following types: the number of days between the observation date and the departure date (1 feature), the boolean variables for day-of-the-week of the observation date (7 features), the complete enumeration of the following 4 values: 1) the minimum price, mean price, and number of quotes from 2) all airlines and from each airline quoting more than 50\% of the observation days 3) for non-stop, one-stop, and two-stop flights, 4) for the current day, previous day, and two days previous. The result is a feature set of 411 variables. For specific details on how these datasets are constructed please consult~\cite{groves2015tist}. The nature of these datasets is heterogeneous with a mixture of several types of variables including boolean variables, prices, and counts.

\noindent \textbf{RF}
The river flow datasets concern the prediction of river network flows for 48 hours in the future at specific locations. The dataset contains data from hourly flow observations for 8 sites in the Mississippi River network in the United States and were obtained from the US National Weather Service. Each row includes the most recent observation for each of the 8 sites as well as time-lagged observations from 6, 12, 18, 24, 36, 48 and 60 hours in the past. In RF1, each site contributes 8 attribute variables to facilitate prediction.
There are a total of 64 variables plus 8 target variables.The RF2 dataset extends the RF1 data by adding precipitation forecast information for each of the 8 sites (expected rainfall reported as discrete values: 0.0, 0.01, 0.25, 1.0 inches). For each observation and gauge site, the precipitation forecast for
6 hour windows up to 48 hours in the future is added (6, 12, 18, 24, 30, 36, 42, and 48 hours).
The two datasets both contain over 1 year of hourly observations ($>$9,000 hours) collected from September 2011 to September 2012. 
The domain is a natural candidate for multi-target regression because there are clear physical relationships between readings in the contiguous river network.

\noindent \textbf{SCM}
The Supply Chain Management datasets are derived from the Trading Agent Competition in Supply Chain Management (TAC SCM) tournament from 2010. The precise methods for data preprocessing and normalization are described in detail by \cite{groves2013improving}. Some benchmark values for prediction accuracy in this domain are available from the TAC SCM Prediction Challenge~\citep{pardoe2008}, these datasets correspond only to the ``Product Future'' prediction type. Each row corresponds to an observation day in the tournament (there are 220 days in each game and 18 tournament games in a tournament). The input variables in this domain are observed prices for a specific tournament day. In addition, 4 time-delayed observations are included for each observed product and component (1,2,4 and 8 days delayed) to facilitate some anticipation of trends going forward. The datasets contain 16 regression targets, each target corresponds to the next day mean price (SCM1D) or mean price for 20-days in the future (SCM20D) for each product in the simulation. Days with no target values are excluded from the datasets (i.e. days with labels that are beyond the end of the game are excluded).

\section{Detailed Experimental Results}
\label{appendix:results}

\subsection{Base Regressor Exploration Results}
\label{appendix:results:st}

\setlength{\LTcapwidth}{\textwidth}
\renewcommand{\tabcolsep}{3pt} 

\begin{longtable}{l *{5}{r}}
\caption{\small{
Detailed results for ST using \textsc{ridge}, \textsc{tree}, \textsc{svr}, \textsc{bag} and \textsc{sgb} as base regressors.
For each dataset, we first report the average $RRMSE$ over all targets ($aRRMSE$), and then the $RRMSE$ per target. In each row, the lowest error is typeset in bold.}}\\
\hline
Dataset & \multicolumn{1}{c}{ST-\textsc{ridge}} & \multicolumn{1}{c}{ST-\textsc{tree}} & \multicolumn{1}{c}{ST-\textsc{svr}} & \multicolumn{1}{c}{ST-\textsc{bag}} & \multicolumn{1}{c}{ST-\textsc{sgb}} \\
\hspace{2 mm} Target &&&&& \\
\hline
\endfirsthead
\multicolumn{4}{c}%
{\tablename\ \thetable\ -- \textit{Continued from previous page}} \\
\endhead
\multicolumn{4}{r}{\textit{Continued on next page}} \\
\endfoot
\endlastfoot
\hline
edm & 0.871 & 0.915 & 0.860 & 0.742 & \textbf{0.706} \\
\hspace{2 mm} dflow & 0.977 & 1.092 & 0.961 & 0.815 & \textbf{0.751} \\
\hspace{2 mm} dgap & 0.764 & 0.737 & 0.760 & 0.669 & \textbf{0.662} \\
\hline
sf1 & 1.130 & 1.127 & \textbf{0.930} & 1.135 & 1.148 \\
\hspace{2 mm} c-class & \textbf{0.987} & 1.045 & 1.001 & 1.017 & 1.026 \\
\hspace{2 mm} m-class & \textbf{1.009} & 1.345 & 1.084 & 1.096 & 1.111 \\
\hspace{2 mm} x-class & 1.394 & 0.991 & \textbf{0.705} & 1.293 & 1.307 \\
\hline
sf2 & 1.485 & 1.010 & \textbf{0.912} & 1.149 & 1.223 \\
\hspace{2 mm} c-class & 0.953 & 0.973 & \textbf{0.943} & 0.980 & 0.980 \\
\hspace{2 mm} m-class & 1.136 & 1.006 & \textbf{0.961} & 1.075 & 1.112 \\
\hspace{2 mm} x-class & 2.365 & 1.050 & \textbf{0.833} & 1.393 & 1.578 \\
\hline
jura & 0.607 & 0.705 & 0.643 & \textbf{0.589} & 0.619 \\
\hspace{2 mm} cd & 0.715 & 0.854 & 0.727 & \textbf{0.711} & 0.750 \\
\hspace{2 mm} co & 0.626 & 0.616 & 0.679 & \textbf{0.543} & 0.550 \\
\hspace{2 mm} cu & \textbf{0.480} & 0.645 & 0.523 & 0.514 & 0.558 \\
\hline
wq & 0.955 & 0.963 & 0.959 & \textbf{0.908} & 0.922 \\
\hspace{2 mm} 25400 & 0.950 & 0.977 & 0.954 & \textbf{0.925} & 0.935 \\
\hspace{2 mm} 29600 & 0.987 & 1.011 & 0.996 & 0.987 & \textbf{0.983} \\
\hspace{2 mm} 30400 & 0.960 & 0.965 & 0.966 & \textbf{0.945} & 0.961 \\
\hspace{2 mm} 33400 & 0.950 & 0.974 & 0.956 & \textbf{0.912} & 0.929 \\
\hspace{2 mm} 17300 & 0.967 & 0.969 & 0.973 & \textbf{0.902} & 0.931 \\
\hspace{2 mm} 19400 & 0.909 & 0.890 & 0.910 & \textbf{0.834} & 0.840 \\
\hspace{2 mm} 34500 & 0.971 & 1.019 & 0.982 & \textbf{0.969} & 0.982 \\
\hspace{2 mm} 38100 & 0.957 & 0.963 & 0.956 & \textbf{0.912} & 0.924 \\
\hspace{2 mm} 49700 & 0.919 & 0.942 & 0.939 & \textbf{0.795} & 0.816 \\
\hspace{2 mm} 50390 & 0.966 & 0.966 & 0.966 & \textbf{0.892} & 0.899 \\
\hspace{2 mm} 55800 & 0.959 & 0.974 & 0.971 & \textbf{0.924} & 0.935 \\
\hspace{2 mm} 57500 & 0.961 & 0.969 & 0.962 & \textbf{0.918} & 0.924 \\
\hspace{2 mm} 59300 & 0.992 & 0.980 & 0.991 & \textbf{0.947} & 0.972 \\
\hspace{2 mm} 37880 & 0.917 & 0.888 & 0.904 & \textbf{0.856} & 0.878 \\
\hline
enb & 0.315 & 0.126 & 0.607 & 0.117 & \textbf{0.114} \\
\hspace{2 mm} y1 & 0.293 & 0.062 & 0.587 & \textbf{0.053} & 0.059 \\
\hspace{2 mm} y2 & 0.336 & 0.189 & 0.627 & 0.180 & \textbf{0.168} \\
\hline
slump & 0.679 & 0.827 & 0.686 & 0.688 & \textbf{0.669} \\
\hspace{2 mm} slump & 0.889 & 0.974 & 0.895 & 0.795 & \textbf{0.744} \\
\hspace{2 mm} flow & 0.774 & 0.857 & 0.755 & 0.742 & \textbf{0.739} \\
\hspace{2 mm} compressi. & \textbf{0.373} & 0.650 & 0.409 & 0.526 & 0.523 \\
\hline
andro & 0.842 & 0.599 & 1.109 & 0.602 & \textbf{0.494} \\
\hspace{2 mm} 1 & 0.801 & 0.663 & 1.038 & 0.515 & \textbf{0.454} \\
\hspace{2 mm} 2 & 0.799 & 0.333 & 0.863 & 0.340 & \textbf{0.219} \\
\hspace{2 mm} 3 & 1.079 & 0.592 & 1.386 & 0.588 & \textbf{0.406} \\
\hspace{2 mm} 4 & 1.079 & 0.522 & 1.458 & 0.530 & \textbf{0.466} \\
\hspace{2 mm} 5 & \textbf{0.676} & 0.704 & 0.985 & 0.809 & 0.686 \\
\hspace{2 mm} 6 & \textbf{0.620} & 0.783 & 0.921 & 0.827 & 0.731 \\
\hline
osales & 0.900 & 0.920 & 0.847 & \textbf{0.748} & 0.807 \\
\hspace{2 mm} m1 & 0.919 & 0.829 & 0.818 & \textbf{0.653} & 0.944 \\
\hspace{2 mm} m2 & 0.916 & 0.868 & 0.824 & \textbf{0.754} & 0.818 \\
\hspace{2 mm} m3 & 0.884 & 0.916 & 0.824 & \textbf{0.786} & 0.800 \\
\hspace{2 mm} m4 & 0.909 & 0.812 & 0.818 & \textbf{0.689} & 0.732 \\
\hspace{2 mm} m5 & 0.991 & 0.917 & 0.848 & \textbf{0.736} & 0.790 \\
\hspace{2 mm} m6 & 0.819 & 0.965 & 0.814 & \textbf{0.696} & 0.711 \\
\hspace{2 mm} m7 & 0.883 & 0.960 & 0.819 & \textbf{0.743} & 0.825 \\
\hspace{2 mm} m8 & 0.911 & 0.870 & 0.841 & \textbf{0.764} & 0.782 \\
\hspace{2 mm} m9 & 0.872 & 1.205 & 0.884 & \textbf{0.812} & 0.915 \\
\hspace{2 mm} m10 & 0.931 & 0.881 & 0.873 & \textbf{0.773} & 0.785 \\
\hspace{2 mm} m11 & 0.845 & 0.841 & 0.872 & 0.749 & \textbf{0.736} \\
\hspace{2 mm} m12 & 0.919 & 0.981 & 0.932 & \textbf{0.821} & 0.846 \\
\hline
scpf & 0.853 & 0.934 & 0.903 & \textbf{0.837} & 0.913 \\
\hspace{2 mm} views & 0.822 & 0.878 & 0.820 & \textbf{0.815} & 0.887 \\
\hspace{2 mm} votes & 0.752 & 0.813 & 0.843 & \textbf{0.720} & 0.754 \\
\hspace{2 mm} comments & 0.985 & 1.111 & 1.045 & \textbf{0.976} & 1.099 \\
\hline
atp1d & 0.412 & 0.473 & 0.432 & \textbf{0.374} & 0.395 \\
\hspace{2 mm} allminpa & 0.476 & 0.652 & 0.491 & 0.482 & \textbf{0.470} \\
\hspace{2 mm} allminp0 & \textbf{0.412} & 0.567 & 0.427 & 0.430 & 0.468 \\
\hspace{2 mm} adlminpa & 0.419 & 0.468 & 0.433 & \textbf{0.416} & 0.428 \\
\hspace{2 mm} acominpa & 0.342 & 0.274 & 0.381 & \textbf{0.242} & 0.273 \\
\hspace{2 mm} aflminpa & 0.472 & 0.647 & 0.466 & 0.471 & \textbf{0.464} \\
\hspace{2 mm} auaminpa & 0.349 & 0.229 & 0.392 & \textbf{0.200} & 0.267 \\
\hline
atp7d & 0.579 & 0.665 & 0.615 & 0.525 & \textbf{0.517} \\
\hspace{2 mm} allminpa & 0.730 & 0.950 & 0.764 & 0.641 & \textbf{0.558} \\
\hspace{2 mm} allminp0 & \textbf{0.575} & 0.811 & 0.595 & 0.670 & 0.635 \\
\hspace{2 mm} adlminpa & 0.566 & 0.699 & 0.577 & 0.546 & \textbf{0.533} \\
\hspace{2 mm} acominpa & 0.437 & 0.355 & 0.509 & \textbf{0.316} & 0.321 \\
\hspace{2 mm} aflminpa & 0.717 & 0.821 & 0.730 & \textbf{0.689} & 0.745 \\
\hspace{2 mm} auaminpa & 0.448 & 0.353 & 0.514 & \textbf{0.286} & 0.313 \\
\hline
oes97 & \textbf{0.477} & 0.709 & 0.579 & 0.525 & 0.644 \\
\hspace{2 mm} 58028 & 0.199 & 0.562 & \textbf{0.183} & 0.336 & 0.462 \\
\hspace{2 mm} 15014 & \textbf{0.282} & 0.510 & 0.292 & 0.358 & 0.554 \\
\hspace{2 mm} 32511 & 0.746 & 0.775 & 1.113 & \textbf{0.717} & 0.754 \\
\hspace{2 mm} 15017 & 0.322 & 0.454 & \textbf{0.316} & 0.365 & 0.380 \\
\hspace{2 mm} 98502 & \textbf{0.602} & 0.815 & 0.896 & 0.665 & 0.682 \\
\hspace{2 mm} 92965 & \textbf{0.595} & 0.961 & 0.766 & 0.648 & 0.766 \\
\hspace{2 mm} 32314 & 0.786 & 0.827 & 0.959 & \textbf{0.614} & 0.837 \\
\hspace{2 mm} 13008 & 0.228 & 0.505 & \textbf{0.217} & 0.339 & 0.517 \\
\hspace{2 mm} 21114 & 0.210 & 0.303 & \textbf{0.209} & 0.307 & 0.395 \\
\hspace{2 mm} 85110 & \textbf{0.448} & 0.754 & 0.770 & 0.567 & 0.669 \\
\hspace{2 mm} 27311 & \textbf{0.573} & 0.905 & 0.652 & 0.601 & 0.786 \\
\hspace{2 mm} 98902 & \textbf{0.457} & 0.850 & 0.461 & 0.536 & 0.689 \\
\hspace{2 mm} 65032 & \textbf{0.470} & 0.754 & 0.484 & 0.552 & 0.580 \\
\hspace{2 mm} 92998 & \textbf{0.615} & 0.828 & 0.617 & 0.644 & 0.701 \\
\hspace{2 mm} 27108 & \textbf{0.544} & 0.700 & 0.669 & 0.580 & 0.700 \\
\hspace{2 mm} 53905 & \textbf{0.553} & 0.839 & 0.665 & 0.569 & 0.825 \\
\hline
oes10 & \textbf{0.369} & 0.621 & 0.427 & 0.420 & 0.569 \\
\hspace{2 mm} 513021 & \textbf{0.368} & 0.646 & 0.481 & 0.438 & 0.416 \\
\hspace{2 mm} 292071 & \textbf{0.367} & 0.599 & 0.413 & 0.385 & 0.404 \\
\hspace{2 mm} 392021 & 0.425 & 0.559 & 0.534 & \textbf{0.412} & 0.492 \\
\hspace{2 mm} 151131 & \textbf{0.346} & 0.655 & 0.363 & 0.430 & 0.649 \\
\hspace{2 mm} 151141 & \textbf{0.355} & 0.502 & 0.434 & 0.436 & 0.455 \\
\hspace{2 mm} 291069 & \textbf{0.576} & 0.971 & 0.646 & 0.616 & 0.876 \\
\hspace{2 mm} 119032 & 0.285 & 0.579 & \textbf{0.280} & 0.370 & 0.490 \\
\hspace{2 mm} 432011 & \textbf{0.321} & 0.589 & 0.426 & 0.392 & 0.487 \\
\hspace{2 mm} 419022 & \textbf{0.522} & 0.858 & 0.636 & 0.644 & 0.686 \\
\hspace{2 mm} 292037 & \textbf{0.325} & 0.459 & 0.376 & 0.335 & 0.412 \\
\hspace{2 mm} 519061 & 0.333 & 0.606 & \textbf{0.331} & 0.395 & 0.426 \\
\hspace{2 mm} 291051 & \textbf{0.180} & 0.347 & 0.192 & 0.265 & 0.404 \\
\hspace{2 mm} 172141 & 0.567 & 0.814 & 0.836 & \textbf{0.494} & 0.856 \\
\hspace{2 mm} 431011 & 0.150 & 0.456 & \textbf{0.139} & 0.269 & 0.616 \\
\hspace{2 mm} 291127 & \textbf{0.386} & 0.674 & 0.412 & 0.462 & 0.564 \\
\hspace{2 mm} 412021 & 0.397 & 0.615 & \textbf{0.338} & 0.377 & 0.878 \\
\hline
rf1 & 0.541 & 0.121 & 0.414 & \textbf{0.097} & 0.230 \\
\hspace{2 mm} chsi2 & 0.334 & 0.077 & 0.214 & \textbf{0.043} & 0.153 \\
\hspace{2 mm} nasi2 & 1.951 & \textbf{0.376} & 1.301 & 0.432 & 0.653 \\
\hspace{2 mm} eadm7 & 0.399 & 0.091 & 0.329 & \textbf{0.055} & 0.182 \\
\hspace{2 mm} sclm7 & 0.600 & 0.128 & 0.367 & \textbf{0.061} & 0.222 \\
\hspace{2 mm} clkm7 & 0.251 & 0.079 & 0.236 & \textbf{0.049} & 0.161 \\
\hspace{2 mm} vali2 & 0.341 & 0.089 & 0.366 & \textbf{0.057} & 0.216 \\
\hspace{2 mm} napm7 & 0.242 & 0.053 & 0.245 & \textbf{0.040} & 0.102 \\
\hspace{2 mm} dldi4 & 0.211 & 0.073 & 0.253 & \textbf{0.039} & 0.151 \\
\hline
rf2 & 0.469 & 0.121 & 0.414 & \textbf{0.102} & 0.237 \\
\hspace{2 mm} chsi2 & 0.228 & 0.078 & 0.214 & \textbf{0.044} & 0.132 \\
\hspace{2 mm} nasi2 & 1.996 & \textbf{0.384} & 1.296 & 0.465 & 0.780 \\
\hspace{2 mm} eadm7 & 0.253 & 0.095 & 0.329 & \textbf{0.055} & 0.167 \\
\hspace{2 mm} sclm7 & 0.294 & 0.105 & 0.367 & \textbf{0.065} & 0.186 \\
\hspace{2 mm} clkm7 & 0.297 & 0.082 & 0.236 & \textbf{0.050} & 0.162 \\
\hspace{2 mm} vali2 & 0.286 & 0.090 & 0.370 & \textbf{0.056} & 0.213 \\
\hspace{2 mm} napm7 & 0.208 & 0.055 & 0.245 & \textbf{0.041} & 0.105 \\
\hspace{2 mm} dldi4 & 0.190 & 0.078 & 0.253 & \textbf{0.040} & 0.153 \\
\hline
scm1d & 0.394 & 0.444 & 0.457 & \textbf{0.348} & 0.393 \\
\hspace{2 mm} lbl & 0.337 & 0.379 & 0.409 & \textbf{0.310} & 0.335 \\
\hspace{2 mm} mtlp2 & 0.350 & 0.401 & 0.436 & \textbf{0.323} & 0.366 \\
\hspace{2 mm} mtlp3 & 0.379 & 0.431 & 0.442 & \textbf{0.333} & 0.388 \\
\hspace{2 mm} mtlp4 & 0.387 & 0.458 & 0.461 & \textbf{0.345} & 0.385 \\
\hspace{2 mm} mtlp5 & 0.454 & 0.508 & 0.530 & \textbf{0.377} & 0.452 \\
\hspace{2 mm} mtlp6 & 0.456 & 0.506 & 0.540 & \textbf{0.376} & 0.437 \\
\hspace{2 mm} mtlp7 & 0.449 & 0.496 & 0.526 & \textbf{0.370} & 0.455 \\
\hspace{2 mm} mtlp8 & 0.451 & 0.523 & 0.497 & \textbf{0.377} & 0.445 \\
\hspace{2 mm} mtlp9 & 0.372 & 0.421 & 0.456 & \textbf{0.341} & 0.378 \\
\hspace{2 mm} mtlp10 & 0.394 & 0.428 & 0.456 & \textbf{0.352} & 0.424 \\
\hspace{2 mm} mtlp11 & 0.377 & 0.417 & 0.445 & \textbf{0.342} & 0.364 \\
\hspace{2 mm} mtlp12 & 0.404 & 0.447 & 0.466 & \textbf{0.366} & 0.411 \\
\hspace{2 mm} mtlp13 & 0.363 & 0.419 & 0.409 & \textbf{0.331} & 0.357 \\
\hspace{2 mm} mtlp14 & 0.394 & 0.472 & 0.432 & \textbf{0.369} & 0.383 \\
\hspace{2 mm} mtlp15 & 0.356 & 0.406 & 0.393 & \textbf{0.330} & 0.352 \\
\hspace{2 mm} mtlp16 & 0.373 & 0.398 & 0.407 & \textbf{0.330} & 0.361 \\
\hline
scm20d & 0.646 & 0.627 & 0.763 & \textbf{0.475} & 0.620 \\
\hspace{2 mm} lbl & 0.561 & 0.562 & 0.678 & \textbf{0.424} & 0.542 \\
\hspace{2 mm} mtlp2a & 0.571 & 0.574 & 0.688 & \textbf{0.425} & 0.542 \\
\hspace{2 mm} mtlp3a & 0.586 & 0.572 & 0.683 & \textbf{0.440} & 0.564 \\
\hspace{2 mm} mtlp4a & 0.614 & 0.612 & 0.730 & \textbf{0.455} & 0.580 \\
\hspace{2 mm} mtlp5a & 0.707 & 0.681 & 0.846 & \textbf{0.493} & 0.697 \\
\hspace{2 mm} mtlp6a & 0.694 & 0.650 & 0.843 & \textbf{0.493} & 0.641 \\
\hspace{2 mm} mtlp7a & 0.691 & 0.644 & 0.833 & \textbf{0.485} & 0.624 \\
\hspace{2 mm} mtlp8a & 0.690 & 0.648 & 0.851 & \textbf{0.500} & 0.655 \\
\hspace{2 mm} mtlp9a & 0.644 & 0.604 & 0.737 & \textbf{0.454} & 0.599 \\
\hspace{2 mm} mtlp10a & 0.660 & 0.630 & 0.753 & \textbf{0.478} & 0.663 \\
\hspace{2 mm} mtlp11a & 0.666 & 0.667 & 0.769 & \textbf{0.498} & 0.632 \\
\hspace{2 mm} mtlp12a & 0.673 & 0.668 & 0.787 & \textbf{0.511} & 0.669 \\
\hspace{2 mm} mtlp13a & 0.651 & 0.634 & 0.751 & \textbf{0.481} & 0.611 \\
\hspace{2 mm} mtlp14a & 0.661 & 0.645 & 0.779 & \textbf{0.501} & 0.683 \\
\hspace{2 mm} mtlp15a & 0.633 & 0.613 & 0.727 & \textbf{0.480} & 0.608 \\
\hspace{2 mm} mtlp16a & 0.636 & 0.635 & 0.754 & \textbf{0.488} & 0.610 \\
\hline
\label{tbl:detailed_st}
\end{longtable}

\subsection{Multi-target Regression Results}
\label{appendix:results:mtr}

\small{
\begin{longtable}{l *{11}{r}}
\caption{\small{
Detailed results for all methods using \textsc{bag} as base regressor in ST, SST, ERC and RLC (detailed results with additional base regressors can be found at \url{http://users.auth.gr/espyromi/mtr/results.zip}). For each dataset, we first report the average $RRMSE$ over all targets ($aRRMSE$), and then the $RRMSE$ per target. In each row, the lowest error is typeset in bold.}}\\
\hline
Dataset & \multicolumn{1}{c}{\rotatebox{90}{ST}} & \multicolumn{1}{c}{\rotatebox{90}{SST$_{true}$}} & \multicolumn{1}{c}{\rotatebox{90}{SST$_{train}$}} & \multicolumn{1}{c}{\rotatebox{90}{SST$_{cv}$}} & \multicolumn{1}{c}{\rotatebox{90}{ERC$_{true}$}} & \multicolumn{1}{c}{\rotatebox{90}{ERC$_{train}$}} & \multicolumn{1}{c}{\rotatebox{90}{ERC$_{cv}$}} & \multicolumn{1}{c}{\rotatebox{90}{MORF}} & \multicolumn{1}{c}{\rotatebox{90}{RLC}} & \multicolumn{1}{c}{\rotatebox{90}{TNR}} & \multicolumn{1}{c}{\rotatebox{90}{Dirty}}\\
\hspace{2 mm} Target &&&&&&&&& \\
\hline
\endfirsthead
\multicolumn{10}{c}%
{\tablename\ \thetable\ -- \textit{Continued from previous page}} \\
\endhead
\multicolumn{10}{r}{\textit{Continued on next page}} \\
\endfoot
\endlastfoot
\hline
edm & 0.742 & 0.747 & 0.743 & 0.740 & 0.743 & 0.742 & 0.741 & \textbf{0.734} & 0.735 & 0.851 & 0.830 \\
\hspace{2 mm} dflow & 0.815 & 0.824 & 0.817 & 0.812 & 0.818 & 0.815 & 0.814 & \textbf{0.775} & 0.801 & 0.932 & 0.900 \\
\hspace{2 mm} dgap & 0.669 & 0.671 & 0.669 & \textbf{0.667} & 0.669 & 0.669 & 0.668 & 0.692 & 0.669 & 0.769 & 0.759 \\
\hline
sf1 & 1.135 & \textbf{0.997} & 1.127 & 1.068 & 1.050 & 1.132 & 1.089 & 1.282 & 1.163 & 1.112 & 1.115 \\
\hspace{2 mm} c-class & 1.017 & 0.991 & 1.037 & 1.001 & 1.000 & 1.020 & 1.007 & 1.035 & 1.019 & 0.974 & \textbf{0.973} \\
\hspace{2 mm} m-class & 1.096 & \textbf{0.918} & 1.137 & 1.005 & 0.992 & 1.112 & 1.033 & 1.212 & 1.130 & 0.998 & 1.016 \\
\hspace{2 mm} x-class & 1.293 & \textbf{1.083} & 1.207 & 1.198 & 1.158 & 1.263 & 1.226 & 1.601 & 1.341 & 1.365 & 1.356 \\
\hline
sf2 & 1.149 & 0.980 & \textbf{0.945} & 1.055 & 1.053 & 1.087 & 1.088 & 1.425 & 1.228 & 1.475 & 1.372 \\
\hspace{2 mm} c-class & 0.980 & 0.968 & 0.975 & 0.964 & 0.973 & 0.977 & 0.964 & 0.996 & 0.985 & 0.948 & \textbf{0.944} \\
\hspace{2 mm} m-class & 1.075 & \textbf{0.983} & 0.994 & 0.992 & 1.022 & 1.018 & 1.030 & 1.160 & 1.080 & 1.118 & 1.103 \\
\hspace{2 mm} x-class & 1.393 & 0.988 & \textbf{0.866} & 1.210 & 1.165 & 1.266 & 1.270 & 2.119 & 1.620 & 2.360 & 2.069 \\
\hline
jura & \textbf{0.589} & 0.594 & 0.592 & 0.591 & 0.591 & 0.590 & 0.590 & 0.597 & 0.596 & 0.608 & 0.610 \\
\hspace{2 mm} cd & 0.711 & 0.715 & 0.715 & 0.713 & 0.712 & 0.713 & 0.712 & \textbf{0.694} & 0.702 & 0.715 & 0.716 \\
\hspace{2 mm} co & 0.543 & 0.553 & 0.544 & 0.543 & 0.546 & \textbf{0.542} & 0.543 & 0.566 & 0.558 & 0.628 & 0.629 \\
\hspace{2 mm} cu & 0.514 & 0.514 & 0.517 & 0.515 & 0.514 & 0.515 & 0.514 & 0.530 & 0.530 & \textbf{0.481} & 0.485 \\
\hline
wq & 0.908 & 0.914 & 0.911 & 0.909 & 0.910 & 0.905 & 0.906 & \textbf{0.899} & 0.902 & 0.962 & 0.961 \\
\hspace{2 mm} 25400 & 0.925 & 0.931 & 0.930 & 0.928 & 0.930 & \textbf{0.921} & 0.927 & 0.924 & 0.922 & 0.952 & 0.956 \\
\hspace{2 mm} 29600 & 0.987 & 0.988 & 0.986 & 0.984 & 0.985 & 0.983 & 0.983 & \textbf{0.976} & 0.979 & 0.994 & 0.995 \\
\hspace{2 mm} 30400 & 0.945 & 0.945 & 0.945 & 0.951 & 0.944 & 0.944 & 0.946 & 0.942 & \textbf{0.937} & 0.968 & 0.964 \\
\hspace{2 mm} 33400 & 0.912 & 0.915 & 0.904 & 0.902 & 0.912 & 0.904 & 0.902 & \textbf{0.893} & 0.904 & 0.959 & 0.957 \\
\hspace{2 mm} 17300 & 0.902 & 0.913 & 0.921 & 0.914 & 0.906 & 0.910 & 0.908 & \textbf{0.895} & 0.903 & 0.974 & 0.973 \\
\hspace{2 mm} 19400 & 0.834 & 0.839 & 0.829 & 0.829 & 0.835 & \textbf{0.827} & 0.829 & 0.828 & 0.832 & 0.912 & 0.907 \\
\hspace{2 mm} 34500 & 0.969 & 0.965 & 0.961 & 0.963 & 0.965 & 0.958 & 0.957 & 0.959 & \textbf{0.957} & 0.981 & 0.976 \\
\hspace{2 mm} 38100 & 0.912 & 0.913 & 0.908 & 0.908 & 0.912 & 0.906 & 0.911 & 0.907 & \textbf{0.904} & 0.967 & 0.967 \\
\hspace{2 mm} 49700 & 0.795 & 0.809 & 0.815 & 0.808 & 0.799 & 0.796 & 0.796 & \textbf{0.793} & 0.793 & 0.927 & 0.936 \\
\hspace{2 mm} 50390 & 0.892 & 0.899 & 0.901 & 0.899 & 0.892 & 0.892 & 0.892 & 0.892 & \textbf{0.884} & 0.988 & 0.961 \\
\hspace{2 mm} 55800 & 0.924 & 0.934 & 0.931 & 0.929 & 0.926 & 0.922 & 0.924 & \textbf{0.903} & 0.916 & 0.963 & 0.973 \\
\hspace{2 mm} 57500 & 0.918 & 0.920 & 0.917 & 0.919 & 0.918 & 0.915 & 0.914 & \textbf{0.896} & 0.907 & 0.965 & 0.964 \\
\hspace{2 mm} 59300 & 0.947 & 0.965 & 0.957 & 0.951 & 0.954 & 0.946 & 0.947 & \textbf{0.931} & 0.941 & 0.995 & 0.994 \\
\hspace{2 mm} 37880 & 0.856 & 0.860 & 0.848 & \textbf{0.847} & 0.857 & 0.852 & 0.849 & 0.851 & 0.851 & 0.923 & 0.927 \\
\hline
enb & 0.117 & 0.145 & 0.123 & 0.121 & 0.125 & 0.117 & \textbf{0.114} & 0.121 & 0.120 & 0.316 & 0.319 \\
\hspace{2 mm} y1 & 0.053 & 0.088 & 0.059 & 0.063 & 0.064 & 0.053 & 0.053 & 0.060 & \textbf{0.053} & 0.295 & 0.297 \\
\hspace{2 mm} y2 & 0.180 & 0.201 & 0.186 & 0.178 & 0.187 & 0.181 & \textbf{0.174} & 0.182 & 0.188 & 0.337 & 0.341 \\
\hline
slump & 0.688 & 0.722 & \textbf{0.666} & 0.695 & 0.701 & 0.669 & 0.689 & 0.694 & 0.690 & 0.681 & 0.684 \\
\hspace{2 mm} slump & 0.795 & 0.857 & \textbf{0.742} & 0.802 & 0.817 & 0.757 & 0.796 & 0.775 & 0.792 & 0.900 & 0.900 \\
\hspace{2 mm} flow & 0.742 & 0.767 & 0.743 & 0.764 & 0.752 & \textbf{0.732} & 0.749 & 0.733 & 0.740 & 0.769 & 0.769 \\
\hspace{2 mm} compressi. & 0.526 & 0.540 & 0.514 & 0.521 & 0.533 & 0.519 & 0.522 & 0.573 & 0.539 & \textbf{0.372} & 0.384 \\
\hline
andro & 0.602 & 0.603 & 0.540 & 0.579 & 0.596 & 0.538 & 0.567 & \textbf{0.510} & 0.570 & 0.803 & 0.889 \\
\hspace{2 mm} 1 & 0.515 & 0.584 & 0.487 & 0.507 & 0.537 & 0.485 & 0.504 & \textbf{0.436} & 0.452 & 0.860 & 0.792 \\
\hspace{2 mm} 2 & 0.340 & 0.349 & 0.340 & 0.336 & 0.344 & 0.339 & 0.332 & 0.403 & \textbf{0.296} & 1.000 & 0.934 \\
\hspace{2 mm} 3 & 0.588 & 0.497 & \textbf{0.404} & 0.518 & 0.540 & 0.435 & 0.475 & 0.499 & 0.594 & 0.833 & 1.009 \\
\hspace{2 mm} 4 & 0.530 & 0.551 & 0.500 & 0.579 & 0.521 & 0.483 & 0.521 & \textbf{0.467} & 0.587 & 0.837 & 1.052 \\
\hspace{2 mm} 5 & 0.809 & 0.843 & 0.842 & 0.800 & 0.819 & 0.758 & 0.788 & \textbf{0.632} & 0.745 & 0.657 & 0.761 \\
\hspace{2 mm} 6 & 0.827 & 0.795 & 0.667 & 0.736 & 0.816 & 0.730 & 0.781 & \textbf{0.622} & 0.747 & 0.629 & 0.782 \\
\hline
osales & 0.748 & 0.751 & 0.709 & 0.726 & 0.728 & \textbf{0.699} & 0.713 & 0.753 & 0.741 & 1.628 & 1.507 \\
\hspace{2 mm} m1 & 0.653 & 0.695 & 0.604 & 0.596 & 0.674 & \textbf{0.590} & 0.596 & 0.676 & 0.657 & 1.881 & 1.737 \\
\hspace{2 mm} m2 & 0.754 & 0.740 & 0.696 & 0.756 & 0.733 & \textbf{0.675} & 0.732 & 0.719 & 0.750 & 1.551 & 1.643 \\
\hspace{2 mm} m3 & 0.786 & 0.809 & 0.779 & 0.825 & \textbf{0.760} & 0.772 & 0.773 & 0.778 & 0.772 & 1.639 & 1.630 \\
\hspace{2 mm} m4 & 0.689 & 0.728 & \textbf{0.636} & 0.660 & 0.696 & 0.643 & 0.659 & 0.736 & 0.676 & 1.598 & 1.640 \\
\hspace{2 mm} m5 & 0.736 & 0.668 & 0.669 & 0.652 & 0.692 & 0.653 & \textbf{0.647} & 0.738 & 0.703 & 2.319 & 1.810 \\
\hspace{2 mm} m6 & 0.696 & 0.763 & 0.725 & 0.735 & 0.704 & \textbf{0.671} & 0.699 & 0.753 & 0.710 & 1.562 & 1.555 \\
\hspace{2 mm} m7 & 0.743 & 0.726 & 0.682 & 0.691 & 0.712 & \textbf{0.673} & 0.679 & 0.768 & 0.738 & 1.572 & 1.322 \\
\hspace{2 mm} m8 & 0.764 & 0.760 & 0.729 & \textbf{0.726} & 0.746 & 0.730 & 0.733 & 0.789 & 0.759 & 1.359 & 1.293 \\
\hspace{2 mm} m9 & 0.812 & 0.767 & 0.755 & 0.803 & 0.747 & \textbf{0.741} & 0.778 & 0.746 & 0.793 & 1.401 & 1.394 \\
\hspace{2 mm} m10 & 0.773 & 0.763 & 0.729 & 0.733 & 0.738 & \textbf{0.726} & 0.732 & 0.770 & 0.776 & 1.905 & 1.471 \\
\hspace{2 mm} m11 & 0.749 & 0.756 & \textbf{0.708} & 0.734 & 0.737 & 0.710 & 0.729 & 0.760 & 0.736 & 1.278 & 1.202 \\
\hspace{2 mm} m12 & 0.821 & 0.843 & 0.799 & 0.801 & \textbf{0.797} & 0.798 & 0.801 & 0.806 & 0.818 & 1.471 & 1.384 \\
\hline
scpf & 0.837 & 0.830 & 0.855 & 0.831 & \textbf{0.812} & 0.821 & 0.830 & 0.833 & 0.835 & 0.899 & 0.877 \\
\hspace{2 mm} views & 0.815 & 0.795 & 0.811 & 0.815 & \textbf{0.787} & 0.789 & 0.810 & 0.808 & 0.814 & 0.855 & 0.857 \\
\hspace{2 mm} votes & 0.720 & 0.750 & 0.759 & 0.718 & 0.718 & 0.724 & 0.719 & \textbf{0.704} & 0.724 & 0.825 & 0.752 \\
\hspace{2 mm} comments & 0.976 & 0.945 & 0.996 & 0.960 & \textbf{0.931} & 0.950 & 0.961 & 0.988 & 0.966 & 1.015 & 1.020 \\
\hline
atp1d & 0.374 & 0.376 & 0.372 & 0.372 & 0.371 & \textbf{0.367} & 0.372 & 0.422 & 0.378 & 0.595 & 0.552 \\
\hspace{2 mm} allminpa & 0.482 & 0.506 & 0.485 & 0.482 & 0.485 & 0.481 & 0.482 & \textbf{0.474} & 0.475 & 0.687 & 0.644 \\
\hspace{2 mm} allminp0 & 0.430 & 0.428 & 0.420 & 0.422 & 0.426 & \textbf{0.416} & 0.428 & 0.436 & 0.434 & 0.687 & 0.628 \\
\hspace{2 mm} adlminpa & 0.416 & \textbf{0.397} & 0.405 & 0.410 & 0.398 & 0.402 & 0.412 & 0.424 & 0.419 & 0.564 & 0.519 \\
\hspace{2 mm} acominpa & 0.242 & 0.245 & \textbf{0.233} & 0.242 & 0.242 & 0.234 & 0.239 & 0.356 & 0.248 & 0.481 & 0.441 \\
\hspace{2 mm} aflminpa & 0.471 & 0.468 & 0.480 & 0.473 & 0.470 & 0.470 & 0.473 & 0.487 & \textbf{0.461} & 0.654 & 0.627 \\
\hspace{2 mm} auaminpa & \textbf{0.200} & 0.210 & 0.207 & 0.202 & 0.205 & 0.201 & 0.200 & 0.356 & 0.232 & 0.497 & 0.453 \\
\hline
atp7d & 0.525 & 0.561 & 0.514 & \textbf{0.507} & 0.534 & 0.509 & 0.512 & 0.551 & 0.529 & 0.786 & 0.668 \\
\hspace{2 mm} allminpa & 0.641 & 0.765 & \textbf{0.619} & 0.629 & 0.696 & 0.619 & 0.626 & 0.636 & 0.673 & 0.828 & 0.761 \\
\hspace{2 mm} allminp0 & 0.670 & 0.698 & 0.672 & 0.653 & 0.675 & 0.666 & 0.660 & \textbf{0.602} & 0.644 & 0.827 & 0.738 \\
\hspace{2 mm} adlminpa & 0.546 & 0.588 & 0.533 & 0.537 & 0.553 & 0.530 & 0.542 & \textbf{0.524} & 0.555 & 0.784 & 0.624 \\
\hspace{2 mm} acominpa & 0.316 & 0.301 & \textbf{0.269} & 0.278 & 0.294 & 0.283 & 0.291 & 0.437 & 0.328 & 0.705 & 0.561 \\
\hspace{2 mm} aflminpa & 0.689 & 0.727 & 0.728 & 0.700 & 0.707 & 0.692 & 0.694 & 0.674 & \textbf{0.669} & 0.852 & 0.745 \\
\hspace{2 mm} auaminpa & 0.286 & 0.288 & 0.264 & \textbf{0.246} & 0.281 & 0.267 & 0.262 & 0.432 & 0.304 & 0.721 & 0.577 \\
\hline
oes97 & 0.525 & 0.526 & 0.526 & 0.524 & 0.525 & 0.525 & 0.524 & 0.549 & \textbf{0.523} & 0.818 & 1.049 \\
\hspace{2 mm} 58028 & 0.336 & 0.337 & 0.335 & 0.337 & 0.337 & 0.335 & 0.336 & \textbf{0.265} & 0.312 & 0.354 & 0.389 \\
\hspace{2 mm} 15014 & 0.358 & 0.359 & 0.365 & \textbf{0.352} & 0.359 & 0.361 & 0.352 & 0.465 & 0.367 & 0.446 & 0.560 \\
\hspace{2 mm} 32511 & 0.717 & 0.715 & 0.715 & \textbf{0.714} & 0.716 & 0.716 & 0.714 & 0.752 & 0.733 & 1.398 & 2.077 \\
\hspace{2 mm} 15017 & 0.365 & \textbf{0.365} & 0.370 & 0.368 & 0.365 & 0.367 & 0.366 & 0.403 & 0.373 & 0.505 & 0.743 \\
\hspace{2 mm} 98502 & 0.665 & 0.675 & 0.666 & 0.665 & 0.671 & 0.666 & 0.665 & 0.715 & \textbf{0.659} & 1.219 & 1.653 \\
\hspace{2 mm} 92965 & 0.648 & 0.651 & 0.648 & 0.647 & 0.649 & 0.647 & 0.647 & 0.698 & \textbf{0.640} & 0.970 & 1.626 \\
\hspace{2 mm} 32314 & 0.614 & \textbf{0.611} & 0.615 & 0.617 & 0.613 & 0.614 & 0.615 & 0.704 & 0.633 & 0.915 & 1.137 \\
\hspace{2 mm} 13008 & 0.339 & 0.339 & 0.340 & 0.337 & 0.339 & 0.340 & 0.338 & \textbf{0.272} & 0.321 & 0.318 & 0.460 \\
\hspace{2 mm} 21114 & 0.307 & 0.307 & 0.307 & 0.308 & 0.307 & 0.307 & 0.307 & 0.291 & \textbf{0.283} & 0.419 & 0.458 \\
\hspace{2 mm} 85110 & 0.567 & 0.567 & 0.575 & 0.568 & 0.567 & 0.569 & 0.566 & 0.636 & \textbf{0.566} & 0.902 & 1.186 \\
\hspace{2 mm} 27311 & 0.601 & 0.598 & 0.595 & 0.595 & 0.599 & 0.598 & 0.598 & \textbf{0.566} & 0.584 & 0.952 & 1.108 \\
\hspace{2 mm} 98902 & 0.536 & 0.537 & 0.537 & 0.536 & 0.536 & 0.536 & 0.535 & 0.543 & \textbf{0.513} & 0.704 & 0.743 \\
\hspace{2 mm} 65032 & 0.552 & 0.552 & 0.554 & 0.550 & 0.552 & 0.553 & 0.551 & \textbf{0.535} & 0.542 & 0.877 & 0.956 \\
\hspace{2 mm} 92998 & 0.644 & 0.653 & \textbf{0.642} & 0.644 & 0.647 & 0.643 & 0.642 & 0.703 & 0.659 & 1.198 & 1.252 \\
\hspace{2 mm} 27108 & 0.580 & 0.582 & \textbf{0.576} & 0.577 & 0.581 & 0.578 & 0.577 & 0.636 & 0.596 & 0.842 & 1.256 \\
\hspace{2 mm} 53905 & \textbf{0.569} & 0.572 & 0.574 & 0.572 & 0.570 & 0.571 & 0.570 & 0.602 & 0.591 & 1.064 & 1.173 \\
\hline
oes10 & 0.420 & 0.421 & 0.420 & 0.421 & 0.420 & 0.420 & 0.420 & 0.452 & \textbf{0.419} & 0.532 & 0.664 \\
\hspace{2 mm} 513021 & 0.438 & \textbf{0.436} & 0.438 & 0.437 & 0.436 & 0.438 & 0.438 & 0.453 & 0.443 & 0.587 & 0.824 \\
\hspace{2 mm} 292071 & 0.385 & 0.387 & 0.383 & \textbf{0.382} & 0.387 & 0.383 & 0.383 & 0.393 & 0.388 & 0.548 & 0.733 \\
\hspace{2 mm} 392021 & 0.412 & 0.413 & 0.413 & 0.414 & 0.413 & 0.412 & 0.412 & \textbf{0.397} & 0.415 & 0.605 & 0.807 \\
\hspace{2 mm} 151131 & 0.430 & 0.431 & 0.431 & 0.431 & 0.431 & 0.431 & 0.430 & 0.457 & \textbf{0.423} & 0.511 & 0.515 \\
\hspace{2 mm} 151141 & \textbf{0.436} & 0.440 & 0.439 & 0.441 & 0.438 & 0.437 & 0.438 & 0.496 & 0.446 & 0.508 & 0.770 \\
\hspace{2 mm} 291069 & 0.616 & 0.618 & 0.617 & 0.616 & 0.617 & 0.617 & 0.616 & 0.650 & \textbf{0.589} & 0.834 & 0.874 \\
\hspace{2 mm} 119032 & 0.370 & 0.370 & 0.369 & 0.371 & 0.370 & 0.368 & 0.370 & 0.402 & 0.384 & \textbf{0.332} & 0.509 \\
\hspace{2 mm} 432011 & 0.392 & 0.393 & 0.392 & 0.392 & 0.392 & 0.392 & 0.392 & 0.397 & \textbf{0.391} & 0.471 & 0.661 \\
\hspace{2 mm} 419022 & 0.644 & 0.644 & 0.644 & 0.644 & 0.644 & 0.645 & 0.644 & 0.654 & \textbf{0.610} & 0.779 & 0.914 \\
\hspace{2 mm} 292037 & \textbf{0.335} & 0.336 & 0.337 & 0.337 & 0.335 & 0.336 & 0.335 & 0.382 & 0.352 & 0.432 & 0.606 \\
\hspace{2 mm} 519061 & \textbf{0.395} & 0.396 & 0.397 & 0.397 & 0.395 & 0.396 & 0.396 & 0.557 & 0.404 & 0.558 & 0.608 \\
\hspace{2 mm} 291051 & 0.265 & 0.266 & 0.267 & 0.266 & 0.265 & 0.265 & 0.264 & 0.266 & 0.262 & \textbf{0.220} & 0.297 \\
\hspace{2 mm} 172141 & \textbf{0.494} & 0.495 & 0.496 & 0.496 & 0.495 & 0.496 & 0.494 & 0.593 & 0.520 & 1.067 & 1.080 \\
\hspace{2 mm} 431011 & 0.269 & 0.268 & 0.267 & 0.271 & 0.268 & 0.268 & 0.269 & 0.227 & 0.240 & 0.166 & \textbf{0.165} \\
\hspace{2 mm} 291127 & 0.462 & 0.460 & 0.453 & 0.456 & 0.461 & 0.456 & 0.458 & \textbf{0.442} & 0.456 & 0.449 & 0.794 \\
\hspace{2 mm} 412021 & 0.377 & \textbf{0.375} & 0.378 & 0.379 & 0.376 & 0.377 & 0.378 & 0.461 & 0.378 & 0.437 & 0.474 \\
\hline
rf1 & 0.097 & 0.113 & 0.094 & 0.094 & 0.101 & 0.091 & \textbf{0.091} & 0.123 & 0.121 & 0.983 & 0.676 \\
\hspace{2 mm} chsi2 & 0.043 & 0.069 & 0.047 & 0.046 & 0.050 & 0.035 & 0.034 & 0.035 & \textbf{0.033} & 0.797 & 0.387 \\
\hspace{2 mm} nasi2 & 0.432 & 0.498 & 0.431 & 0.431 & 0.454 & 0.430 & \textbf{0.430} & 0.650 & 0.684 & 1.946 & 2.610 \\
\hspace{2 mm} eadm7 & 0.055 & 0.050 & 0.043 & 0.040 & 0.047 & 0.041 & 0.040 & \textbf{0.039} & 0.042 & 1.019 & 0.524 \\
\hspace{2 mm} sclm7 & 0.061 & 0.069 & 0.049 & 0.048 & 0.060 & 0.048 & \textbf{0.047} & 0.068 & 0.048 & 1.503 & 0.834 \\
\hspace{2 mm} clkm7 & 0.049 & 0.048 & 0.041 & 0.041 & 0.045 & 0.040 & 0.041 & \textbf{0.031} & 0.041 & 0.587 & 0.256 \\
\hspace{2 mm} vali2 & 0.057 & 0.074 & 0.056 & 0.057 & 0.069 & 0.055 & 0.055 & \textbf{0.037} & 0.046 & 0.571 & 0.320 \\
\hspace{2 mm} napm7 & 0.040 & 0.051 & 0.044 & 0.047 & 0.045 & 0.039 & 0.041 & 0.097 & \textbf{0.038} & 0.909 & 0.270 \\
\hspace{2 mm} dldi4 & 0.039 & 0.046 & 0.038 & 0.038 & 0.041 & 0.037 & 0.038 & \textbf{0.029} & 0.031 & 0.534 & 0.208 \\
\hline
rf2 & 0.102 & 0.123 & 0.100 & 0.097 & 0.109 & 0.096 & \textbf{0.095} & 0.148 & 0.130 & 1.103 & 0.586 \\
\hspace{2 mm} chsi2 & 0.044 & 0.077 & 0.042 & 0.046 & 0.059 & 0.035 & \textbf{0.034} & 0.066 & 0.038 & 0.737 & 0.258 \\
\hspace{2 mm} nasi2 & 0.465 & 0.534 & 0.470 & \textbf{0.456} & 0.482 & 0.461 & 0.461 & 0.587 & 0.747 & 3.143 & 2.597 \\
\hspace{2 mm} eadm7 & 0.055 & 0.055 & 0.047 & \textbf{0.040} & 0.052 & 0.043 & 0.041 & 0.071 & 0.046 & 0.737 & 0.319 \\
\hspace{2 mm} sclm7 & 0.065 & 0.089 & 0.056 & 0.050 & 0.074 & 0.053 & \textbf{0.049} & 0.119 & 0.051 & 0.970 & 0.370 \\
\hspace{2 mm} clkm7 & 0.050 & 0.049 & \textbf{0.041} & 0.041 & 0.048 & 0.042 & 0.042 & 0.054 & 0.042 & 0.891 & 0.420 \\
\hspace{2 mm} vali2 & 0.056 & 0.074 & 0.056 & 0.054 & 0.069 & 0.054 & 0.053 & 0.067 & \textbf{0.047} & 0.956 & 0.270 \\
\hspace{2 mm} napm7 & 0.041 & 0.055 & 0.051 & 0.049 & 0.047 & 0.044 & 0.043 & 0.166 & \textbf{0.039} & 0.617 & 0.273 \\
\hspace{2 mm} dldi4 & 0.040 & 0.049 & 0.039 & 0.039 & 0.042 & 0.039 & 0.039 & 0.053 & \textbf{0.032} & 0.770 & 0.180 \\
\hline
scm1d & 0.348 & 0.360 & 0.340 & 0.336 & 0.353 & 0.332 & \textbf{0.330} & 0.352 & 0.345 & 0.437 & 0.399 \\
\hspace{2 mm} lbl & 0.310 & 0.312 & 0.300 & 0.296 & 0.310 & 0.294 & \textbf{0.294} & 0.308 & 0.306 & 0.378 & 0.341 \\
\hspace{2 mm} mtlp2 & 0.323 & 0.334 & 0.317 & 0.313 & 0.329 & 0.309 & \textbf{0.308} & 0.327 & 0.317 & 0.408 & 0.353 \\
\hspace{2 mm} mtlp3 & 0.333 & 0.351 & 0.325 & 0.322 & 0.342 & 0.319 & \textbf{0.315} & 0.328 & 0.327 & 0.429 & 0.384 \\
\hspace{2 mm} mtlp4 & 0.345 & 0.362 & 0.338 & 0.334 & 0.359 & 0.330 & \textbf{0.325} & 0.343 & 0.342 & 0.442 & 0.393 \\
\hspace{2 mm} mtlp5 & 0.377 & 0.404 & 0.366 & 0.365 & 0.395 & 0.357 & \textbf{0.349} & 0.391 & 0.377 & 0.502 & 0.461 \\
\hspace{2 mm} mtlp6 & 0.376 & 0.392 & 0.357 & 0.350 & 0.381 & 0.356 & \textbf{0.347} & 0.399 & 0.382 & 0.509 & 0.462 \\
\hspace{2 mm} mtlp7 & 0.370 & 0.380 & 0.350 & 0.343 & 0.376 & 0.346 & \textbf{0.338} & 0.383 & 0.369 & 0.497 & 0.459 \\
\hspace{2 mm} mtlp8 & 0.377 & 0.393 & 0.362 & 0.356 & 0.390 & 0.353 & \textbf{0.345} & 0.391 & 0.378 & 0.504 & 0.459 \\
\hspace{2 mm} mtlp9 & 0.341 & 0.349 & 0.333 & 0.330 & 0.344 & \textbf{0.323} & 0.325 & 0.340 & 0.332 & 0.411 & 0.376 \\
\hspace{2 mm} mtlp10 & 0.352 & 0.359 & 0.350 & 0.344 & 0.351 & \textbf{0.339} & 0.341 & 0.348 & 0.348 & 0.434 & 0.398 \\
\hspace{2 mm} mtlp11 & 0.342 & 0.348 & 0.331 & 0.329 & 0.342 & \textbf{0.327} & 0.329 & 0.346 & 0.336 & 0.410 & 0.384 \\
\hspace{2 mm} mtlp12 & 0.366 & 0.366 & 0.355 & 0.355 & 0.362 & \textbf{0.350} & 0.354 & 0.363 & 0.364 & 0.441 & 0.412 \\
\hspace{2 mm} mtlp13 & 0.331 & 0.348 & 0.334 & 0.326 & 0.337 & 0.323 & \textbf{0.322} & 0.345 & 0.326 & 0.399 & 0.367 \\
\hspace{2 mm} mtlp14 & 0.369 & 0.374 & 0.363 & 0.360 & 0.367 & \textbf{0.356} & 0.358 & 0.361 & 0.358 & 0.438 & 0.398 \\
\hspace{2 mm} mtlp15 & 0.330 & 0.337 & 0.321 & 0.319 & 0.327 & \textbf{0.314} & 0.315 & 0.330 & 0.322 & 0.389 & 0.361 \\
\hspace{2 mm} mtlp16 & 0.330 & 0.346 & 0.333 & 0.329 & 0.334 & 0.322 & \textbf{0.322} & 0.335 & 0.332 & 0.405 & 0.380 \\
\hline
scm20d & 0.475 & 0.493 & 0.431 & 0.413 & 0.497 & 0.415 & \textbf{0.394} & 0.443 & 0.472 & 0.655 & 0.658 \\
\hspace{2 mm} lbl & 0.424 & 0.441 & 0.386 & 0.368 & 0.448 & 0.372 & \textbf{0.356} & 0.393 & 0.423 & 0.569 & 0.561 \\
\hspace{2 mm} mtlp2a & 0.425 & 0.442 & 0.384 & 0.365 & 0.455 & 0.369 & \textbf{0.352} & 0.402 & 0.429 & 0.581 & 0.573 \\
\hspace{2 mm} mtlp3a & 0.440 & 0.453 & 0.402 & 0.387 & 0.464 & 0.385 & \textbf{0.363} & 0.415 & 0.437 & 0.593 & 0.587 \\
\hspace{2 mm} mtlp4a & 0.455 & 0.471 & 0.410 & 0.396 & 0.485 & 0.397 & \textbf{0.374} & 0.426 & 0.453 & 0.622 & 0.618 \\
\hspace{2 mm} mtlp5a & 0.493 & 0.535 & 0.459 & 0.441 & 0.532 & 0.432 & \textbf{0.413} & 0.467 & 0.489 & 0.711 & 0.737 \\
\hspace{2 mm} mtlp6a & 0.493 & 0.512 & 0.450 & 0.430 & 0.504 & 0.438 & \textbf{0.424} & 0.466 & 0.488 & 0.701 & 0.728 \\
\hspace{2 mm} mtlp7a & 0.485 & 0.499 & 0.440 & 0.422 & 0.502 & 0.422 & \textbf{0.404} & 0.450 & 0.482 & 0.698 & 0.718 \\
\hspace{2 mm} mtlp8a & 0.500 & 0.504 & 0.447 & 0.431 & 0.513 & 0.426 & \textbf{0.407} & 0.454 & 0.493 & 0.696 & 0.725 \\
\hspace{2 mm} mtlp9a & 0.454 & 0.481 & 0.421 & 0.402 & 0.489 & 0.404 & \textbf{0.382} & 0.437 & 0.458 & 0.650 & 0.647 \\
\hspace{2 mm} mtlp10a & 0.478 & 0.497 & 0.436 & \textbf{0.413} & 0.498 & 0.435 & 0.418 & 0.454 & 0.489 & 0.668 & 0.666 \\
\hspace{2 mm} mtlp11a & 0.498 & 0.502 & 0.428 & 0.407 & 0.505 & 0.426 & \textbf{0.402} & 0.449 & 0.490 & 0.671 & 0.670 \\
\hspace{2 mm} mtlp12a & 0.511 & 0.525 & 0.458 & 0.444 & 0.520 & 0.449 & \textbf{0.429} & 0.473 & 0.511 & 0.680 & 0.681 \\
\hspace{2 mm} mtlp13a & 0.481 & 0.515 & 0.453 & 0.431 & 0.516 & 0.426 & \textbf{0.400} & 0.450 & 0.471 & 0.663 & 0.658 \\
\hspace{2 mm} mtlp14a & 0.501 & 0.517 & 0.455 & 0.440 & 0.520 & 0.436 & \textbf{0.411} & 0.465 & 0.492 & 0.675 & 0.668 \\
\hspace{2 mm} mtlp15a & 0.480 & 0.496 & 0.432 & 0.411 & 0.503 & 0.409 & \textbf{0.384} & 0.439 & 0.472 & 0.648 & 0.638 \\
\hspace{2 mm} mtlp16a & 0.488 & 0.496 & 0.434 & 0.414 & 0.504 & 0.411 & \textbf{0.386} & 0.445 & 0.481 & 0.647 & 0.646 \\
\hline
\label{tbl:detailed_mt}
\end{longtable}
}

\renewcommand{\tabcolsep}{6pt} 

\end{document}